\documentclass[journal]{IEEEtran}
\usepackage{times}
\usepackage{soul}
\usepackage{bm}
\usepackage{url}
\usepackage{xcolor}
\definecolor{mydarkgreen}{rgb}{0,0.6,0}
\definecolor{mydarkblue}{rgb}{0,0,0.6}
\usepackage[colorlinks,
linkcolor=mydarkgreen,
citecolor=mydarkblue,
urlcolor=black]{hyperref}
\usepackage[utf8]{inputenc}
\usepackage[small]{caption}
\usepackage{graphicx}
\usepackage{subfigure}
\usepackage{amsmath,amssymb,amsfonts}
\usepackage{amsthm}
\usepackage{mathrsfs}
\usepackage{booktabs}
\usepackage{algorithm}
\usepackage{algorithmic}
\usepackage{multirow}
\newcommand{\httpsurl}[1]{\href{https://#1}{\nolinkurl{#1}}}

\newtheorem{theorem}{Theorem}
\newtheorem{definition}{Definition}
\newtheorem{definition*}{Problem}
\newtheorem{remark}{Remark}

\ifCLASSINFOpdf
\else
\fi
\begin{document}
\title{Learning from a Complementary-label Source Domain: Theory and Algorithms}

\author{Yiyang Zhang$^\dagger$, Feng Liu$^\dagger$,~\IEEEmembership{Student Member,~IEEE},~Zhen Fang,\\Bo Yuan,~
        Guangquan~Zhang,
        and~Jie~Lu$^*$,~\IEEEmembership{Fellow,~IEEE}% <-this % stops a space
%\thanks{Yiyang Zhang, Bo Yuan are with Shenzhen International Graduate School, Tsinghua University, Shenzhen, P.R. China. Feng Liu, Zhen Fang, Guangquan Zhang and Jie Lu are with the Centre for Artificial Intelligence, Faulty of Engineering and Information Technology, University of Technology Sydney, Sydney,
%NSW, 2007, Australia, e-mail: \{Feng.Liu-2, zhen.fang\}@student.uts.edu.au, \{Guangquan.Zhang; Jie.Lu\}@uts.edu.au.}% <-this % stops a space
% \thanks{J. Doe and J. Doe are with Anonymous University.}% <-this % stops a space
% \thanks{Manuscript received April 19, 2005; revised August 26, 2015.}
\thanks{Yiyang Zhang is with Centre for Artificial Intelligence, Faulty of Engineering and Information Technology, University of Technology Sydney, Sydney, NSW, 2007, Australia, and Shenzhen International Graduate School, Tsinghua University, Shenzhen, P.R. China (e-mail: zhangyiy18@mails.tsinghua.edu.cn; yiyang.zhang@uts.edu.au).

 Feng Liu, Zhen Fang, Guangquan Zhang, and Jie Lu are with the Centre for Artificial Intelligence, Faulty of Engineering and Information Technology, University of Technology Sydney, Sydney, NSW, 2007, Australia (e-mail: feng.liu@uts.edu.au; zhen.fang@student.uts.edu.au; guangquan.zhang@uts.edu.au; jie.lu@uts.edu.au).

Bo Yuan is with Shenzhen International Graduate School, Tsinghua University, Shenzhen, P.R. China (e-mail: yuanb@sz.tsinghua.edu.cn).

$^\dagger$Equal contribution. $^*$Corresponding author.}}

\maketitle

\begin{abstract}
% Collecting true-label source data is costly and thus a critical bottleneck in promoting \emph{unsupervised domain adaptation} (UDA) to more fields. In contrast to the standard UDA paradigm where the classifier for the {target domain} is trained with massive \emph{true-label} data from the {source domain} and unlabeled data from the target domain, we propose a novel setting that the source domain is composed of \emph{complementary-label} data---we name it \emph{budget-friendly UDA} (BFUDA), and a theoretical bound for BFUDA is first proved. A complementary label specifies a class that a pattern does not belong to, hence collecting complementary labels would be less laborious than collecting true labels. We consider two cases of BFUDA, one is that the source domain only contains complementary-label data (completely complementary unsupervised domain adaptation, CC-UDA), and the other is that the source domain has plenty of complementary-label data and a small amount of true-label data (partly complementary unsupervised domain adaptation, PC-UDA). To this end, \emph{\underline{c}omplementary \underline{l}abel advers\underline{ari}al \underline{net}work} (CLARINET) is proposed to solve BFUDA problem. CLARINET maintains two deep networks simultaneously, where one focuses on classifying complementary-label source data and the other takes care of source-to-target distributional adaptation. Experiments show that CLARINET significantly outperforms a series of competent baselines.

In \emph{unsupervised domain adaptation} (UDA), a classifier for the {target domain} is trained with massive \emph{true-label} data from the {source domain} and unlabeled data from the target domain. However, collecting fully-true-label data in the source domain is high-cost and sometimes impossible. Compared to the true labels, a complementary label specifies a class that a pattern does not belong to, hence collecting complementary labels would be less laborious than collecting true labels. Thus, in this paper, we propose a novel setting that the source domain is composed of \emph{complementary-label} data, and a theoretical bound for it is first proved. We consider two cases of this setting, one is that the source domain only contains complementary-label data (completely complementary unsupervised domain adaptation, CC-UDA), and the other is that the source domain has plenty of complementary-label data and a small amount of true-label data (partly complementary unsupervised domain adaptation, PC-UDA). To this end, a \emph{\underline{c}omplementary \underline{l}abel advers\underline{ari}al \underline{net}work} (CLARINET) is proposed to solve CC-UDA and PC-UDA problems. CLARINET maintains two deep networks simultaneously, where one focuses on classifying complementary-label source data and the other takes care of source-to-target distributional adaptation. Experiments show that CLARINET significantly outperforms a series of competent baselines on handwritten-digits-recognition and objects-recognition tasks.
\end{abstract}


% \fi
\IEEEpeerreviewmaketitle

\section{Introduction}

\IEEEPARstart{D} {omain} Adaptation (DA) aims to train a target-domain classifier with data in source and target domains \cite{luo2018transferring,xiao2014feature,zhang2013domain}. Based on the availability of data in the target domain (e.g., fully-labeled, partially-labeled and unlabeled), DA is divided into three categories: supervised DA \cite{motiian2017unified,sukhija2016,tzeng2015simultaneous}, semi-supervised DA \cite{DBLP:conf/ijcai/Yan0WMTW18,DBLP:journals/tnn/MehrkanoonS18,islam2019semi} and \emph{unsupervised DA} (UDA) \cite{DBLP:journals/tnn/WeiKG19,saito2017asymmetric,cao2018unsupervised}. In practical applications, UDA is more challenging and promising than the other two as the labeled target domain data are not needed \cite{deng2019cluster,zhao2019geometry,DBLP}.
%\IEEEPARstart{D} {omain} Adaptation (DA) aims to train a target-domain classifier with data in source and target domains \cite{tzeng2017adversarial,yan2017learning,luo2018transferring,xiao2014feature,zhang2013domain}. Based on the availability of data in the target domain (e.g., fully-labeled, partially-labeled and unlabeled), DA is divided into three categories: supervised DA \cite{motiian2017unified,sukhija2016,tzeng2015simultaneous}, semi-supervised DA \cite{DBLP:conf/ijcai/Yan0WMTW18,ao2017fast,islam2019semi} and \emph{unsupervised DA} (UDA) \cite{bousmalis2017unsupervised,saito2017asymmetric,cao2018unsupervised}. In practice, UDA methods have been applied to solve many real-world problems \cite{deng2019cluster,zhao2019geometry,DBLP}.

UDA methods train a target-domain classifier with massive true-label data from the source domain (true-label source data) and unlabeled data from the target domain (unlabeled target data).  Existing works in the literature can be roughly categorised into the following three groups: integral-probability-metrics based UDA \cite{long2015learning,DBLP:conf/icml/LongZ0J17}; adversarial-training based UDA \cite{ganin2016domain,long2018conditional}; and causality-based UDA \cite{Gong2016CTC,Zhang15MSDA_causal}. Since adversarial-training based UDA methods extract better domain-invariant representations via deep networks, they usually have good target-domain accuracy \cite{sankaranarayanan2018generate}.

\begin{figure}[!tp]
    \centering
    \includegraphics[width=0.4\textwidth]{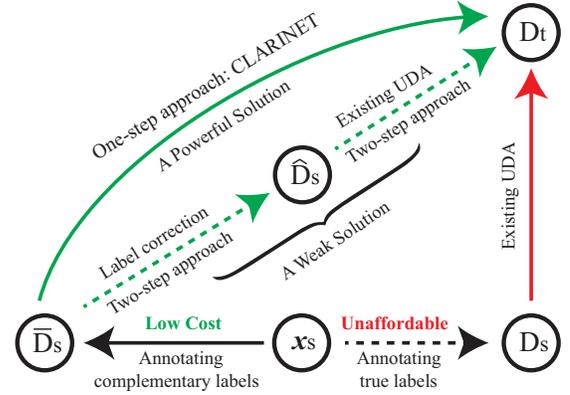}
	\caption{\footnotesize Complementary-label based UDA. The red line denotes that UDA methods transfer knowledge from $D_s$ (true-label source data) to $D_t$ (unlabeled target data). However, acquiring fully-true-label source data is \emph{costly} and \emph{unaffordable} (black dash line, $\mathbf{x}_s \rightarrow {D}_s$, $\mathbf{x}_s$ means unlabeled source data). This brings complementary-label based UDA, namely transferring knowledge from $\overline{D}_s$ (complementary-label source data) to $D_t$. It is much less costly to collect complementary-label source data (black line, required by our setting) than collecting the true-label one (black dash line, required by UDA). To handle complementary-label based UDA, a weak solution is a two-step approach (green dash line), which sequentially combines complementary-label learning methods ($\overline{D}_s \rightarrow \hat{D}_s$, label correction) and existing UDA methods ($\hat{D}_s \rightarrow D_t$). This paper proposes a one-step approach called \emph{\underline{c}omplementary \underline{l}abel advers\underline{ari}al \underline{net}work} (CLARINET, green line, $\overline{D}_s \rightarrow D_t$ directly).}
	\label{fig: our_solution}
	\vspace{-1em}
\end{figure}

However, the success of UDA still highly relies on the scale of true-label source data (black dash line in Figure~\ref{fig: our_solution}). Namely, the target-domain accuracy of a UDA method (e.g., CDAN) decays when the scale of true-label source data decreases and we prove this phenomenon in the experiment section. Hence, massive true-label source data are inevitably required by UDA methods, which is very expensive and even prohibitive. 

\begin{figure*}[!tp]
	\begin{center}
		{\includegraphics[width=0.8\textwidth]{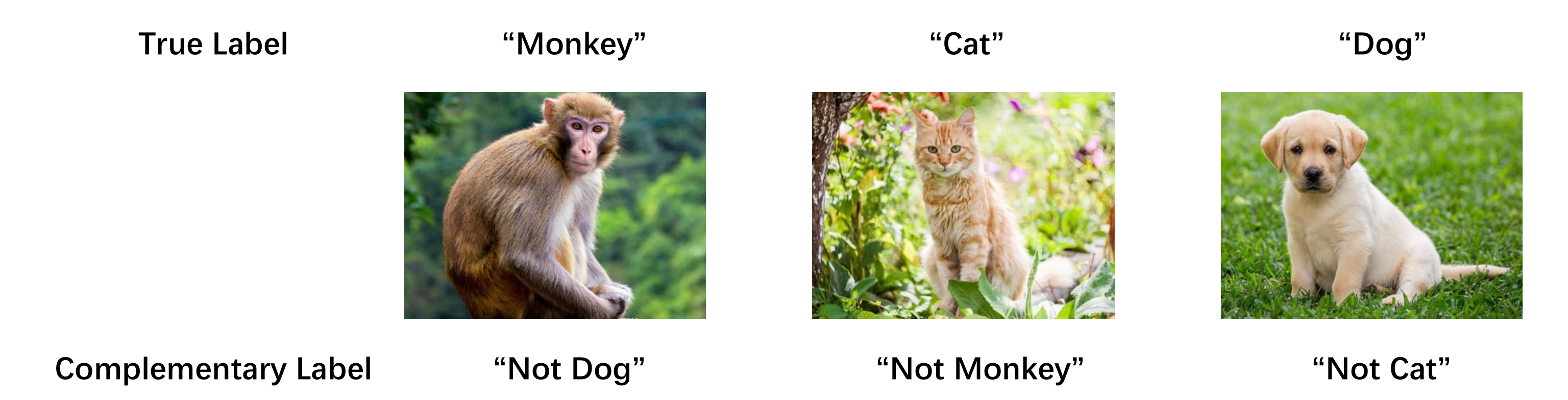}}
		\end{center}
		\vspace{-1em}
		\caption{True label (top) versus complementary label (bottom).}
		\label{fig: complementary}
	\vspace{-1em}
\end{figure*}

While determining the correct label from many candidates is laborious, choosing one of the incorrect labels (i.e., complementary labels),  e.g. labeling a cat as ``Not Monkey" (as shown in Figure~\ref{fig: complementary}), would be much easier and quicker, thus less costly, especially when we have many candidates \cite{ishida2017learning}. For example, suppose that we need to annotate the labels of a bunch of animal images from $1,000$ candidates. One strategy is to ask crowd-workers to choose the true labels from $1,000$ candidates, while the other is to judge the correctness of a label randomly given by the system from the candidates. Apparently, the cost of the second strategy is much lower \cite{pmlr-v97-ishida19a,Feng2020MultiCLL}. 

This brings us a novel setting, complementary-label based UDA, which aims to transfer knowledge from complementary-label source data to unlabeled target data (Figure~\ref{fig: our_solution}). Compared to ordinary UDA, we can greatly save the labeling cost by annotating complementary labels in the source domain rather than annotating true labels  \cite{ishida2017learning,yu2018learning}. Please note, existing UDA methods cannot handle complementary-label based UDA, as they require fully-true-label source data \cite{saito2017asymmetric,cao2018unsupervised} or at least $20\%$ true-label source data \cite{liu2019butterfly,TCL_long}. 

In our previous conference paper \cite{mine}, we consider using completely complementary-label data in the source domain while actually we could also get a small amount of true labels when collecting complementary labels \cite{ishida2017learning}. Therefore, the previous work was flawed as it did not make good use of the existing true-label data. Furthermore, experiments were conducted only on some digit datasets and a thorough learning bound was not provided. Aiming at these defects, in this work, we consider a generalized and completed version of the complementary-label based UDA problem setting.
%In our previous conference paper \cite{mine}, we consider BFUDA setting in the case of using completely complementary-label data in the source domain and conduct experiments only on some digit datasets. We show that distributional adaptation could be effectively achieved  from complementary-label source data to unlabeled target data. On this basis, we extend our previous paper to consider a generalized and completed version of BFUDA problem setting. 
%In addressing the challenge of BFUDA, a previous work \cite{mine} consider using completely complementary-label data in source domain. In that paper, experiments conducted on several digit datasets confirm  that distributional adaptation could be effectively achieved  from complementary-label source data to unlabeled target data. On this basis, some improvements are made in our paper, to consider a more generalized and completed BFUDA problem setting .

A straightforward but weak solution to complementary-label based UDA is a two-step approach, which sequentially combines \emph{complementary-label learning} methods and existing UDA methods (green dash line in Figure~\ref{fig: our_solution})\footnote{We implement this two-step approach and take it as a baseline.}. Complementary-label learning methods are used to assign pseudo labels for complementary-label source data. Then, we can train a target-domain classifier with pseudo-label source data and unlabeled target data using existing UDA methods. Nevertheless, pseudo-label source data contain noise, which may cause poor domain-adaptation performance of this two-step approach \cite{liu2019butterfly}. 

Therefore, we propose a powerful one-step solution,  \emph{\underline{c}omplementary \underline{l}abel advers\underline{ari}al \underline{net}work} (CLARINET). It maintains two deep networks trained by adversarial way simultaneously, where one can accurately classify complementary-label source data, and the other can discriminate source and target domains. Since Long et. al. \cite{long2018conditional} and Song et. al. \cite{song2009hilbert} have shown that the multimodal structures of distributions can only be captured sufficiently by the cross-covariance dependency between the features and classes (i.e., true labels), we set the input of domain discriminator $D$ as the outer product of feature representation (e.g., $\bm{g}_s$ in Figure~\ref{fig: CLARINET}) and mapped classifier prediction (e.g., $\bm{T(f_s)}$ in Figure~\ref{fig: CLARINET}).

%Due to the nature of complementary-label classification, predicted probability of each class (i.e., each element of $\bm{f_s}$, Figure~\ref{fig: CLARINET}) is relatively close.
% Namely, it is hard to find significant predictive preference from the classifier predictions. 
%According to \cite{song2009hilbert}, this kind of predicted probabilities could not provide sufficient information to capture the multimodal structure of distributions. To fix it, we propose a sharpening function $T$ to make the classifier prediction represent its choice well. In this way, we can take full advantage of classifier predictions and effectively align distributions of two domains. As a result, CLARINET has a good adaptation performance.
Due to the nature of complementary-label classification, the predicted probability of each class (i.e., each element of $\bm{f_s}$, Figure~\ref{fig: CLARINET}) is relatively close. According to \cite{song2009hilbert}, this kind of predicted probabilities could not provide sufficient information to capture the multimodal structure of distributions. To fix it, we add a sharpening function $T$ to make the predicted probabilities more scattered (i.e., $\bm{T(f_s)}$, Figure~\ref{fig: CLARINET}) than previous ones (i.e., $\bm{f_s}$, Figure~\ref{fig: CLARINET}). By doing so, the mapped classifier predictions can better indicate their choice. In this way, we can take full advantage of classifier predictions and effectively align distributions of two domains. Our ablation study (see Table~\ref{tab:ablation}) verifies that 
the sharpening function $T$ indeed helps improve the target-domain accuracy.
%sharpened

%Without specifically stated, when we refer to BFUDA, we mainly refer to the case of CC-UDA.

We conduct experiments on $7$ complementary-label based UDA tasks and compare CLARINET with a series of competent baselines. Empirical results demonstrated that CLARINET effectively transfers knowledge from complementary-label source data to unlabeled target data and is superior to all baselines. We also show that the target-domain accuracy of CLARINET will increase if a small amount of true-label source data are available. To make up for the defects of previous conference paper \cite{mine}, the main contributions of this paper are summarized as follows.

%\begin{itemize}
  %\item [1)] 
  %We bring a novel setting {\em budget-friendly UDA} (BFUDA), which aims to use complementary-label source data instead of true-label source data. It can significantly save the labeling cost and thus promote DA to more fields.      
  %\item [2)]
  %We proposed a one-step solution CLARINET to solve BFUDA problems, both CC-UDA tasks and PC-UDA tasks. We also implement several two-step methods and demonstrate the outperformance of CLARINET. In addition, we further show that CLARINET could obtain better transfer result when a few true label exists.
  %also handle the knowledge transfer when few true label exists. 
 % \item [3)]
 % We design and add a sharpening function $T$ in the CLARINET, which helps to capture the multimodal structures of distributions on basis of the characteristics of complementary-label learning. We further show in an ablation study that the sharpening function $T$ plays an important role to improve the target-domain accuracy. 
  %according to the characteristics of complementary label learning. The sharpening function $T$ could scatter the classifier predictions, making them express the predicted goal clearly. 
%\end{itemize}

\begin{itemize}
  \item [1)] 
  We present a generalized version of the complementary-label based UDA. This paper considers two cases of complementary-label based UDA, one is that the source domain only contains complementary-label data (completely complementary unsupervised domain adaptation, CC-UDA), and the other is that the source domain also contains a small amount of true-label data (partly complementary unsupervised domain adaptation, PC-UDA).       
  \item [2)]
  We provide a thorough theoretical analysis of the expected target-domain risk of our approach, presenting a learning bound of complementary-label based UDA.
  \item [3)]
  Apart from the handwritten digit datasets, we also conduct experiments on more complex image datasets, proving the applicability of complementary-label based UDA. 
\end{itemize}

This paper is organized as follows. Section \uppercase\expandafter{\romannumeral2} reviews the works related to domain adaptation, complementary-label learning, and low-cost unsupervised domain adaptation. Section \uppercase\expandafter{\romannumeral3} introduces the problem setting and prove a learning bound of this problem. Section \uppercase\expandafter{\romannumeral4} introduces a straightforward but weak two-step approach to complementary-label based UDA. The proposed powerful one-step solution is shown in Section \uppercase\expandafter{\romannumeral5}. Experimental results and analyses are provided in Section \uppercase\expandafter{\romannumeral6}. Finally, Section \uppercase\expandafter{\romannumeral7} concludes this paper.

\section{Related Works}

In this section, we discuss previous works that are most related to our work, and highlight our differences from them. We mainly review some related works about domain adaptation, complementary-label learning and low-cost unsupervised domain adaptation.

\subsection{Domain Adaptation}

%With the emergence and development of the deep network, machine learning technologies have already achieved significant success in many areas including classification \cite{he2016deep}, regression \cite{held2016learning} and clustering \cite{xie2016unsupervised}. The success is depend on sufficient true-label data to train the models. However, many machine learning methods work well only under a common assumption: the training and test data are drawn from the same feature space and the same distribution \cite{pan2009survey}. When the distribution changes, most models need to be rebuilt from scratch using newly collected training data. There are two shortcomings of doing so, marking data costs much and retraining a model needs a lot of time. As human beings could take advantage of existing knowledge to learn new things, transfer learning \cite{lu2015transfer} has been studied so that machines can also be able to transfer knowledge. Domain adaptation is one of the important part of the transfer knowledge field.

Domain adaptation generalizes a learner across different domains by matching the distributions of source and target domains. It has wide application in computer vision \cite{gopalan2011domain,gong2012geodesic,hoffman2014lsda} and natural language processing \cite{collobert2011natural,DBLP:conf/icml/GlorotBB11}, etc. Previous domain adaptation methods in the shallow regime either try to bridge the source and target by learning invariant feature representations or estimating instance importances using labeled source data and unlabeled target data \cite{huang2007correcting,pan2010domain}. Later, it is confirmed that deep learning methods formed by the composition of multiple non-linear transformations yield abstract and ultimately useful representations \cite{bengio2013repre}. Besides, the learned deep representations to some extent are general and are transferable to similar tasks \cite{DBLP:conf/nips/YosinskiCBL14}. Hence, deep neural networks have been explored for domain adaptation. 

Concurrently, multiple methods of matching the feature distributions in the source and the target domains have been proposed for unsupervised domain adaptation. The first category learns domain invariant features by minimizing a distance between distributions, such as Maximum Mean Discrepancy (MMD) \cite{Liu2020DMMD}. In DAN \cite{long2015learning}, Long et.al. minimize the marginal distributions of two domains by multi-kernel MMD (MK-MMD) metric. An alternative way of learning domain invariant features in UDA is inspired by the Generative Adversarial Networks (GANs). By confusing a domain classifier (or discriminator), the deep networks can explore non-discriminative representations. The adversarial-training based UDA methods always try to play a two-player minimax game. DANN \cite{bousmalis2017unsupervised} employs a gradient reversal layer to realize the minimax optimation. In \cite{long2018conditional}, Long et.al. propose conditional domain adversarial network (CDAN) which conditions the models on discriminative information conveyed in the classifier
predictions. Some works study the UDA problem from a causal
point of view where they consider the label $Y$ is the cause for feature representation $X$. In \cite{Gong2016CTC}, Gong et. al. aim to extract conditional transferable components whose conditional distribution is invariant after proper location-scale (LS) transformations.

However, the aforementioned methods all based on the true-label source domain data, which require high labeling costs. In our work, we propose a new setting which use complementary-label source data instead of true-label source data, which significantly save the labeling cost.
%Existing works in the literature can be roughly categorised into the following two groups: integral-probability-metrics based UDA, like DAN \cite{long2015learning}; and adversarial-training based UDA, like DANN \cite{ganin2016domain} and CDAN \cite{long2018conditional}. However, the aforementioned methods all based on true-label source domain, which requires high labeling costs. In our work, we propose a new setting BFUDA which use complementary-label source data instead of true-label source data, which significantly save the labeling cost.
%Many deep methods seek an explicit feature space transformation that would map source distribution into the target one.

\subsection{Complementary-label Learning}

Complementary-label learning (shown in Figure \ref{fig: complementary}) is one type of weak supervision learning approaches, which is first proposed by Ishida et. al. \cite{ishida2017learning}. They gave theoretical analysis with a statistical consistency guarantee to show classification risk can be recovered only from complementary-label data. Nevertheless, they require the complementary label must be chosen in an unbiased way and allow only one-versus-all and pairwise comparison multi-class loss functions with certain non-convex binary losses. Namely softmax cross-entropy loss, which is the most popular loss in deep learning, could not be used to solve the problem. 

Later, Yu et. al. \cite{yu2018learning} extend the problem setting to where complementary label could be chosen in biased way with the assumption that a small set of easily distinguishable true-label data are available in practice. In their point of view, due to humans are biased toward their own experience, it is unrealistic to guarantee the complementary label is chosen in an unbiased way. For example, if an annotator is more familiar with one class than with another, she is more likely to employ the more familiar one as a complementary label. They solve the problem by employing the forward loss correction technique to adjust the learning objective, but limiting the loss function to softmax cross-entropy loss. They theoretically ensure that the classifier learned with complementary labels converges to the optimal one learned with true labels. 

Recently, Ishida et. al. propose a new unbiased risk estimator \cite{pmlr-v97-ishida19a} under the unbiased label chosen assumption. They make any loss functions available for use and have no implicit assumptions on the classifier, namely the estimator could be used for arbitrary models and losses, including softmax cross-entropy loss. They further investigate correction schemes to make complementary label learning practical and demonstrate the performance. Thus in our paper, we take advantage of this estimator for the source domain classification and generalize it to the unsupervised domain adaptation field. 

\subsection{Low-cost Unsupervised Domain Adaptation}

Unsupervised domain adaptation with low cost source data has recently attracted attention. For instance, in \cite{liu2019butterfly}, Liu et. al. consider the situation that the labeled data in the source domain come from amateur annotators or the Internet \cite{lee2018cleannet,DBLP:journals/pami/SchroffCZ11}. As in the wild, acquiring a large amount of perfectly clean labeled data in the source domain is high-cost and sometimes impossible. They name the problem as {\em wildy unsupervised domain adaptation} (abbreviated as WUDA), which aims to transfer knowledge from noisy labeled data in the source domain to unlabeled target data. They show that WUDA ruins all UDA methods if taking no care of label noise in the source domain and propose a Butterfly framework, a powerful and efficient solution to WUDA.
%We consider another way to save the labeling cost by using complementary label data and propose a new setting named as BFUDA. It aims to transfer knowledge from complementary-label data in source domain to unlabeled target data. In the following sections, we will introduce the BFUDA and explain how to address the BFUDA tasks. 

Long et. al. consider the weakly-supervised domain adaptation, where the source domain with noises in labels, features, or both could be tolerated \cite{TCL_long}. Label noise refers to incorrect labels of images due to errors in manual annotation, and feature noise refers to low-quality pixels of images, which may come from blur, overlap, occlusion, or corruption etc. They present a Transferable Curriculum Learning (TCL) approach, extending from curriculum learning and adversarial learning. The TCL model aims to be robust to both sample noises and distribution shift by employing a curriculum which could tell whether a sample is easy and transferable.

In \cite{mine}, we consider another way to save the labeling cost by using completely complementary-label data in the source domain and prove that  distributional adaptation can be effectively realized from complementary-label source data to unlabeled target data. In this paper, we consider two cases of using complementary-label data in the source domain and prove that we could use a small amount of true-label data to improve the transfer result. Besides, as shown in \cite{ishida2017learning}, we can obtain true-label data and complementary-label data simultaneously so that getting a small amount of true-label data is guaranteed to be low-cost. Furthermore, we provide an analysis of the expected target-domain risk of our approach. In the following sections, we will introduce the complementary-label based UDA and explain how to address such tasks.

\section{Complementary-label Based Unsupervised Domain Adaptation}

In this section, we propose a novel problem setting, complementary-label based UDA, and prove a learning bound for it. Then, we show how it brings benefits to domain adaptation field.

\subsection{Problem Setting}

In complementary-label based UDA, we aim to realize distributional adaptation from complementary-label source data to unlabeled target data. We first consider the situation that there are only complementary-label data in the source domain, namely completely \emph{complementary unsupervised domain adaptation} (CC-UDA). Let $\mathcal{X}\subset \mathbb{R}^d$ be a feature (input) space and $\mathcal{Y}:=\{\mathbf{y}_1,...,\mathbf{y}_c,...,\mathbf{y}_K\}$ be a label (output) space, where $\mathbf{y}_c$ is the one-hot vector for label $c$. A \textit{domain} is defined as follows.
\begin{definition}[Domains for CC-UDA]\label{d3}Given random variables $X_s, X_t \in \mathcal{X}$, $Y_s, \overline{Y}_s, {Y}_t \in \mathcal{Y}$, the {source} and {target domains} are joint distributions $P(X_s, \overline{Y}_s)$ and $P(X_t, {Y}_t)$, where the joint distributions $P(X_s, Y_s)\neq P(X_t, {Y}_t)$ and $P(\overline{Y}_s=\mathbf{y}_c|Y_s=\mathbf{y}_c)=0$ for all $\mathbf{y}_c\in \mathcal{Y}$.
\end{definition}
Then, we propose CC-UDA problem as follows.
\begin{definition*}[CC-UDA]
Given independent and identically distributed (\textit{i.i.d.})~labeled samples $\overline{D}_s=\{(\mathbf{x}_s^i,\overline{\mathbf{y}}_s^i)\}^{\overline{n}_s}_{i=1}$ drawn from the source domain $P(X_s, \overline{Y}_s)$ and \textit{i.i.d.}~unlabeled samples $D_t=\{\mathbf{x}_t^i\}^{n_t}_{i=1}$ drawn from the target marginal distribution $P(X_t)$, {the aim} of CC-UDA is to train a classifier  $F_t:\mathcal{X}\rightarrow \mathcal{Y}$ with $\overline{D}_s$ and $D_t$ such that
$F_t$ can accurately classify target data drawn from $P(X_t)$.
\end{definition*}
It is clear that it is impossible to design a
suitable learning procedure without any assumptions on $P(X_s, \overline{Y}_s)$. In this paper, we use the assumption for unbiased
complementary learning proposed by \cite{ishida2017learning,pmlr-v97-ishida19a}:

\begin{equation}
\label{assumption}
    P(\overline{Y}_s=\mathbf{y}_k|X_s)=\frac{1}{K-1}\sum_{c=1, c\neq k}^K P({Y}_s=\mathbf{y}_c|X_s),
\end{equation}
for all $k, c \in \{1,...,K\}$ and $c\neq k$. This unbiased assumption indicates that the selection of complementary labels for samples is with equal probability. 

Ishida et. al. \cite{ishida2017learning} proposed an efficient way to collect labels through crowdsourcing: we choose one of the classes randomly and ask crowdworkers whether a pattern belongs to the chosen class or not. Then the chosen class is treated as true label if the answer is yes; otherwise, the chosen class is regarded as complementary label. Such a yes/no question is much easier and quicker than selecting the correct class from the list of all candidate classes. In addition, we could guarantee that the complementary-label data gotten through this way are under unbiased assumption in Eq.~\eqref{assumption}. 

As we can obtain true-label data and complementary-label data simultaneously, we also consider the problem that the source domain contains a few true-label data. We name this problem as \emph{partly complementary-label unsupervised domain adaptation} (PC-UDA).

\begin{definition*}[PC-UDA]
Given i.i.d labeled samples $\overline{D}_s=\{(\mathbf{x}_s^i,\overline{\mathbf{y}}_s^i)\}^{\overline{n}_s}_{i=1}$ drawn from the domain $P(X_s, \overline{Y}_s)$, ${D}_s=\{(\mathbf{x}_s^i,{\mathbf{y}}_s^i)\}_{i=\overline{n}_s+1}^{\overline{n}_s+n_s}$ drawn from the domain $P(X_s, {Y}_s)$, and i.i.d unlabeled samples $D_t=\{\mathbf{x}_t^i\}^{n_t}_{i=1}$ drawn from the target marginal distribution $P(X_t)$, {the aim} is to find a target classifier  $F_t:\mathcal{X}\rightarrow \mathcal{Y}$ such that
 $F_t$ \textit{classifies target samples into the correct classes.}
 \end{definition*}
 
It is actually a more common situation to have a small number of true-label data. If we leverage both kind of labeled data properly, we could obtain a more accurate classifier. Ishida et. al. \cite{ishida2017learning} have demonstrated that the usefulness of combining true-label and complementary-label data in classification problem. We will further show that in unsupervised domain adaptation field, we could also use both true-label and complementary-label source data to realize knowledge transfer and utilize the true-label data to improve the result.

\subsection{Learning Bound of complementary-label based UDA}
\label{Theory}

A learning bound of complementary-label based UDA is presented in this subsection. We could prove that we can limit the risk in the target domain. Practitioner may safely skip it.

If given a feature transformation:
\begin{equation}
\begin{split}
    {G}: \mathcal{X}&\rightarrow {\mathcal{X}}_{G}:={G}({\mathcal{X}})
    \\ \mathbf{x} &\rightarrow \mathbf{x}_{{G}}:={G}(\mathbf{x}),
\end{split}
\end{equation}
then the induced distributions related to $P_{X_s}$ and $P_{X_t}$ are
\begin{equation}
\begin{split}
    {G}_{\#}P_{X_s} &:= P( {G}(X_s) );
    \\{G}_{\#}P_{X_t} &:= P({G}(X_t)).
\end{split}
\end{equation}

Following the notations in \cite{DBLP:conf/icml/0002LLJ19}, consider a multi-class classification task with a {\textit{hypothesis space}} $\mathcal{H}_{G}$ of the classifiers
\begin{equation}\label{hyp}
\begin{split}
    ~~~~~~~{F}:~\mathcal{X}_{G}&\rightarrow \mathcal{Y}\\
    {\mathbf{x}}&\rightarrow [C_1({\mathbf{x}}),...,C_{K}({\mathbf{x}})]^T.
    \end{split}
\end{equation}

Let
\begin{equation}
\begin{split}
  \ell: \mathbb{R}^{K}\times\mathbb{R}^{K}&\rightarrow \mathbb{R}_{\geq 0}
  \\ (\mathbf{y}, \tilde{\mathbf{y}} )&\rightarrow \ell(\mathbf{y}, \tilde{\mathbf{y}} ),
  \end{split}
\end{equation}
be the loss function. For convenience, we also require $\ell$ satisfying the following conditions in theoretical part:
\\
1. $\ell$ is symmetric and satisfies triangle inequality;\\
2. $ \ell(\mathbf{y}, \tilde{\mathbf{y}} )=0$ iff $\mathbf{y}= \tilde{\mathbf{y}}$;\\
3. $ \ell(\mathbf{y}, \tilde{\mathbf{y}} )\equiv 1$ if $\mathbf{y}\neq \tilde{\mathbf{y}}$ and $\mathbf{y}, \tilde{\mathbf{y}}$ are one-hot vectors.

We can check many losses satisfying the above conditions such as $0$-$1$ loss $1_{\mathbf{y}\neq \tilde{\mathbf{y}}}$ and $\ell_2$ loss $\frac{1}{2}\|\mathbf{y}- \tilde{\mathbf{y}}\|^2_2$.
 The complementary risk for ${F}\circ {G}$ with respect to $\ell$
over $P({{X}_s,\overline{Y}_s})$ is
\begin{equation*}
    {L}_{\overline{s}}({F}\circ {G})=\mathbb{E}\ell({F}\circ {G}(X_s),\overline{Y}_s).
\end{equation*}

The risks for the decision function ${F}\circ {G}$ with respect to loss $\ell$
over implicit distribution $P({{X}_s,{Y}_s}), P({{X}_t,{Y}_t})$ are:
\begin{equation*}
\begin{split}
& L_s({F}\circ {G})=\mathbb{E}\ell({F}\circ {G}(X_s),{Y}_s),
\\&L_t({F}\circ {G})=\mathbb{E}\ell({F}\circ {G}(X_t),{Y}_t).
\end{split}
\end{equation*}

In this paper, we propose a tighter distance named tensor discrepancy distance. The tensor discrepancy distance can future math  the pseudo conditional distributions.

We consider the following tensor mapping:
\begin{equation}
\begin{split}
    {\otimes_{{F}}}: {\mathcal{X}}_G&\rightarrow {\mathcal{X}}_{G} \otimes \mathcal{Y}_t
    \\ \mathbf{x}_{G}&\rightarrow \mathbf{x}_{G}\otimes {F}(\mathbf{x}_{G}).
    \end{split}
\end{equation}

Then we induce two importance distributions:
\begin{equation}
\begin{split}
    {\otimes_{{F}}}_{\#}P_{X_s} &:= P( {\otimes_{{F}}}({G}(X_s)) );
    \\{\otimes_{{F}}}_{\#}P_{X_t}&:= P({\otimes_{{F}}}({G}(X_t))).
\end{split}
\end{equation}

Using $\mathcal{H}_{G}$, we reconstruct a new hypothetical set:
\begin{equation}
\begin{split}
    &~~~~\Delta_{{F},{G}}:= \{\delta_{\overline{{F}}}:{\mathcal{X}}_{G} \otimes \mathcal{Y}^t\rightarrow \mathbb{R}: \overline{{F}} \in \mathcal{H}_{G} \},
    \end{split}
\end{equation}
where $\delta_{\overline{{F}}}(\mathbf{x}_{G}\otimes \mathbf{y})=|\otimes_{F}(\mathbf{x}_{G})-\otimes_{\overline{{F}}}(\mathbf{x}_{G})|$.
Then the distance between ${\otimes_{{F}}}_{\#}P_{X_s}$ and ${\otimes_{{F}}}_{\#}P_{X_t}$ is:
\begin{equation}\label{eq:d}
\begin{split}
    &~~~d^{\ell}_{\Delta_{{F},{G}}}({\otimes_{{F}}}_{\#}P_{X_s},{\otimes_{{F}}}_{\#}P_{X_t})\\&=\sup_{\delta\in \Delta_{{F},{G}}}\Big| \underset{{{\mathbf{z}}\sim \otimes_{{F}}}_{\#}P_{X_s}}{\mathbb{E}}{{\rm sgn}\circ \delta(\mathbf{z})}-\underset{{{\mathbf{z}}\sim \otimes_{{F}}}_{\#}P_{X_t}}{\mathbb{E}}{{\rm sgn}\circ\delta(\mathbf{z})}\Big|,
    \end{split}
\end{equation}
where ${\rm sgn}$ is the sign function.

It is easy to prove that under the conditions (1)-(3) for loss $\ell$ and for any ${{F}}\in \mathcal{H}_{G}$, we have
\begin{equation}
\begin{split}
   & d_{\Delta_{{F},{G}}}^{\ell}({\otimes_{{F}}}_{\#}P_{X_s},{\otimes_{{F}}}_{\#}P_{X_t})\leq
   d^{\ell}_{{\mathcal{H}_{G}}}({{G}}_{\#}P_{X_s},{{G}}_{\#}P_{X_t}),
   \end{split}
\end{equation}
where $d^{\ell}_{{\mathcal{H}_{G}}}$ is the distribution discrepancy defined in \cite{fang2019open,ghifary2017scatter}.
Then, we introduce our main theorem as follows.
\begin{theorem}\label{Mainthe}
Given a loss function $\ell$ satisfying conditions 1-3 and  a hypothesis $\mathcal{H}_{G} \subset \{{F}:\mathcal{X}_{G}\rightarrow \mathcal{Y}\}$, then under unbiased assumption, for any ${F}\in \mathcal{H}_{G}$,  we have 
\begin{equation*}
\begin{split}
L_t({F}\circ {G})&\leq  \overline{L}_s({F}\circ {G}) +\Lambda+ d_{\Delta_{{F},{G}}}^{\ell}({\otimes_{{F}}}_{\#}P_{X_s},{\otimes_{{F}}}_{\#}P_{X_t}),
 \end{split}
\end{equation*}
where $
\overline{L}_s({F}\circ {G}):=\sum_{k=1}^K\int_{\mathcal{X}} \ell({F}\circ {G}(\mathbf{x}),k) {\rm d}P_{{X}_s}-(K-1) {L}_{\overline{s}}({F}\circ {G})$, $P_{X_s}, P_{X_t}$ are source and target marginal distributions, $\Lambda={\min}_{{F}\in \mathcal{H}_{G}}  ~R_s({F}\circ {G})+R_t({F}\circ {G})$.

\label{theorem1}
\end{theorem}
\begin{proof}
Firstly, we prove that $L_s({F}\circ {G})=\overline{L}_s({F}\circ {G})$. To prove it, we investigate the connection between $L_s({F}\circ {G})$ and ${L}_{\overline{s}}({F}\circ {G})$ under unbiased assumption in Eq.~\eqref{assumption}.
Given $K\times K$ matrix $Q$ whose diagonal elements are $0$ and other elements are $1/K$, we represent the unbiased assumption by \begin{equation}
    \overline{\bm {\bm \eta}}=Q{{\bm \eta}},
\end{equation}
where $\overline{{\bm \eta}}=[P(\overline{Y}_s=\mathbf{y}_1|X_s),...,P(\overline{Y}_s=\mathbf{y}_K|X_s)]^T$ and ${{\bm \eta}}=[P({Y}_s=\mathbf{y}_1|X_s),...,P({Y}_s=\mathbf{y}_K|X_s)]^T$.
Note that $Q$ has inverse matrix $Q^{-1}$ whose diagonal elements are $-(K-2)$ and other elements are $1$. Thus, we have that

\begin{equation}\label{mainequation}
     Q^{-1}\overline{{\bm \eta}}={{\bm \eta}}.
\end{equation}

According to Eq.~\eqref{mainequation}, we have
    $P(Y_s=\mathbf{y}_k|X_s) = 1-(K-1)P(\overline{Y}_s=\mathbf{y}_k|X_s)$,
which implies that
\begin{equation}\label{main}
\begin{split}
    L_s({F}\circ {G})&=\sum_{k=1}^K\int_{\mathcal{X}} \ell({F}\circ {G}(\mathbf{x}),k) {\rm d}P_{{X}_s}\\
    &-(K-1) {L}_{\overline{s}}({F}\circ {G}).
    \end{split}
\end{equation}
Hence, $L_s({F}\circ {G})=\overline{L}_s({F}\circ {G})$. The empirical form of Eq. \eqref{main} is known as complementary-label loss (see Eq.~\eqref{E2}).

%Combing Eq.~\eqref{main} with the domain adaptation bound presented in \cite{ben2010theory}
%\begin{equation*}
%\begin{split}
%L_t({\bm h})&\leq   L_s({\bm h}) +\frac{1}{2}d_{\mathcal{H}}^{\ell}(P_{{X}_s},P_{{X}_t})+\Lambda,
 %\end{split}
%\end{equation*}
%we prove this theorem.

%In this bound, the first term, i.e., Eq. \eqref{main}, is the source classification error based on complementary-label source data. 
%The second term is the distribution discrepancy distance between two domains and $\Lambda$ is the difference in labeling functions across the two domains. The empirical form of Eq. \eqref{main} is known as complementary-label loss (see Eq.~\eqref{E2}).

Next we will prove that
\begin{small}
\begin{equation*}
\begin{split}
L_t({F}\circ {G})-{L}_s({F}\circ {G}) &\leq  \Lambda+ d_{\Delta_{{F},{G}}}^{\ell}({\otimes_{{F}}}_{\#}P_{X_s},{\otimes_{{F}}}_{\#}P_{X_t}).
 \end{split}
\end{equation*}
\end{small}
As if it is true, combined with $L_s({F}\circ {G})=\overline{L}_s({F}\circ {G})$, we could easily prove the theorem. It is clearly that 
\begin{small}
\begin{equation}
\begin{split}
    &L_t({F}\circ {G})-{L}_s({F}\circ {G})\\ =~&\int_{\mathcal{X}\times\mathcal{Y}_t}\ell({F}\circ {G}(\mathbf{x}),\mathbf{y}) {\rm d}P_{{X}_t Y_t}-\int_{\mathcal{X}\times\mathcal{Y}_s}\ell({F}\circ {G}(\mathbf{x}),\mathbf{y}) {\rm d}P_{{X}_s Y_s}\\
    \leq~&L_t(\widetilde{F}\circ {G})+\int_{\mathcal{X}\times\mathcal{Y}^t}\ell({F}\circ {G}(\mathbf{x}),\widetilde{F}\circ {G}(\mathbf{x})) {\rm d}P_{{X}_t Y_t}\\
    +~&L_s(\widetilde{F}\circ {G})-\int_{\mathcal{X}\times\mathcal{Y}_s}\ell({F}\circ {G}(\mathbf{x}),\widetilde{F}\circ {G}(\mathbf{x})) {\rm d}P_{{X}_s Y_s},
\end{split}
\label{eq:d0}
\end{equation}
\end{small}where $\widetilde{F}$ is any function from $\mathcal{H}_{G}$. According to conditions 1-3, we have that

\begin{equation}
    \begin{split}
       \underset{{{\mathbf{z}}\sim \otimes_{{F}}}_{\#}P_{X_s}}{\mathbb{E}}{{\rm sgn}\circ \delta_{\widetilde{F}}(\mathbf{z})}=&\int {{\rm sgn}\circ\delta_{\widetilde{F}}(\mathbf{z})}{\rm d}\otimes_{{{F}} {\#}}P_{X_s}\\
       =&\int_{\mathcal{X}} {|{F}\circ {G}(\mathbf{x})-\widetilde{F}\circ {G}(\mathbf{x})|}{\rm d}P_{{X}_s}\\
       =&\int_{\mathcal{X}} \ell{({F}\circ {G}(\mathbf{x}),\widetilde{F}\circ {G}(\mathbf{x}))}{\rm d}P_{{X}_s},
    \end{split}
\end{equation}
similarly,
\begin{equation}
    \begin{split}
       \underset{{{\mathbf{z}}\sim \otimes_{{F}}}_{\#}P_{X_t}}{\mathbb{E}}{{\rm sgn}\circ \delta_{\widetilde{F}}(\mathbf{z})}=\int_{\mathcal{X}} \ell{({F}\circ {G}(\mathbf{x}),\widetilde{F}\circ {G}(\mathbf{x}))}{\rm d}P_{{X}_t},
    \end{split}
\end{equation}
hence, according to the definition of Eq. \eqref{eq:d}, we have
\begin{equation}
    \begin{split}
         &~~~~~d^{\ell}_{\Delta_{{F},{G}}}({\otimes_{{F}}}_{\#}P_{X_s},{\otimes_{{F}}}_{\#}P_{X_t})\\&=\sup_{\widetilde{F},{G}\in \Delta_{{F},{G}}}\Big|\int_{\mathcal{X}} \ell{({F}\circ {G}(\mathbf{x}),\widetilde{F}\circ {G}(\mathbf{x}))}{\rm d}P_{{X}_s}\\
         &~~~~~~~~~~~~~~-\int_{\mathcal{X}} \ell{({F}\circ {G}(\mathbf{x}),\widetilde{F}\circ {G}(\mathbf{x}))}{\rm d}P_{{X}_t}\Big|.   
    \end{split}
    \label{eq:d2}
\end{equation}
Combining Eq. \eqref{eq:d0} and Eq. \eqref{eq:d2}, we have 
\begin{small}
\begin{equation}
\begin{split}
    &L_t({F}\circ {G})-{L}_s({F}\circ {G})\\ 
    \leq~&\min(L_t(\widetilde{F}\circ {G})+L_s(\widetilde{F}\circ {G}))+d^{\ell}_{\Delta_{{F},{G}}}({\otimes_{{F}}}_{\#}P_{X_s},{\otimes_{{F}}}_{\#}P_{X_t})\\
    =~&\Lambda+ d_{\Delta_{{F},{G}}}^{\ell}({\otimes_{{F}}}_{\#}P_{X_s},{\otimes_{{F}}}_{\#}P_{X_t}).
\end{split}
\label{eq:cd}
\end{equation}
\end{small}
Hence, we prove this theorem.
\end{proof}

\subsection{Benefits to DA Field}
Collecting true-label data is always expensive in the real world. Thus, learning from less expensive data \cite{kumar2017semi,DBLP:journals/tnn/FrenayV14,DBLP:journals/pami/LiZ15,sakai} has been extensively studied in machine learning field, including label-noise leaning \cite{han2018co,han2018masking,DBLP:journals/tnn/MiaoCXGLS16}, pairwise/triple-wise
constraints learning \cite{xing2003distance,DBLP:journals/tnn/CovoesHG13,gneighbourhood}, positive-unlabeled learning \cite{du2014analysis,niu2016theoretical,DBLP:journals/tnn/GongLYT19}, complementary-label learning \cite{ishida2017learning,pmlr-v97-ishida19a,yu2018learning} and so on. Among all these research directions, obtaining complementary labels is a cost-effective option. As described in the previous works mentioned above, compared with choosing the true class out of many candidate classes precisely, collecting complementary labels is obviously much easier and less costly. In addition, a classifier trained with complementary-label data is equivalent to a classifier trained with true-label data as shown in \cite{pmlr-v97-ishida19a}.

Actually in the field of domain adaptation, the high cost of true-label data is also an important issue. At present, the success of DA still highly relies on the scale of true-label source data, which is a critical bottleneck.  Under low cost limitation, it is unrealistic to obtain enough true-label source data and thus cannot achieve a good distribution adaptation result. For the same cost, we can get multiple times more complementary-label data than the true-label data. In addition, the  adaptation scenario is limited to some commonly used datasets, e.g. handwritten digit datasets, as they have sufficient true labels to support distributional adaptation. This fact makes it difficult to generalize domain adaptation to more real-world scenarios where it is needed. Thus if we can reduce the labeling cost in the source domain, for example, by using complementary-label data to replace true-label data (complementary-label based UDA), we can promote domain adaptation to more fields. 

Due to existing UDA methods require at least $20\%$ true-label source data \cite{TCL_long}, they cannot handle complementary-label based UDA problem. To address the problem, we introduce a two-step approach, straightforward but weak solution, and then propose a powerful one-step solution, CLARINET.

\begin{algorithm}[tb]
\small
\caption{Two-step Approach for CC-UDA Tasks}
\label{alg:two}
\textbf{Input}: $\overline{D}_s=\{(\mathbf{x}_s^i,\overline{\mathbf{y}}_s^i)\}^{\overline{n}_s}_{i=1}$, $D_t=\{\mathbf{x}_t^i\}^{n_t}_{i=1}$.\\
\textbf{Output}: the target-domain classifier.
\begin{algorithmic}[1] %[1] enables line numbers
%\STATE \textbf{Initialize} $\theta_{G\circ{F}}$ and $\theta_D$;
\STATE \textbf{Train} a classifier $C$ using $\overline{D}_s=\{(\mathbf{x}_s^i,\overline{\mathbf{y}}_s^i)\}^{\overline{n}_s}_{i=1}$ based on the complementary-label learning algorithm.
\STATE \textbf{Use} $C$ to pseudo-label $\overline{D}_s=\{\mathbf{x}_s^i\}^{\overline{n}_s}_{i=1}$, namely generate pseudo-label source domain data $\hat{D}_s=\{(\mathbf{x}_s^i,\hat{\mathbf{y}}_s^i)\}^{\hat{n}_s}_{i=1}$. 
\STATE \textbf{Apply} normal UDA methods on $\hat{D}_s=\{(\mathbf{x}_s^i,\hat{\mathbf{y}}_s^i)\}^{\hat{n}_s}_{i=1}$ and $D_t=\{\mathbf{x}_t^i\}^{n_t}_{i=1}$ to train a target-domain classifier.
\end{algorithmic}
\end{algorithm}

\begin{figure*}[!tp]
	\begin{center}
		{\includegraphics[width=0.65\textwidth]{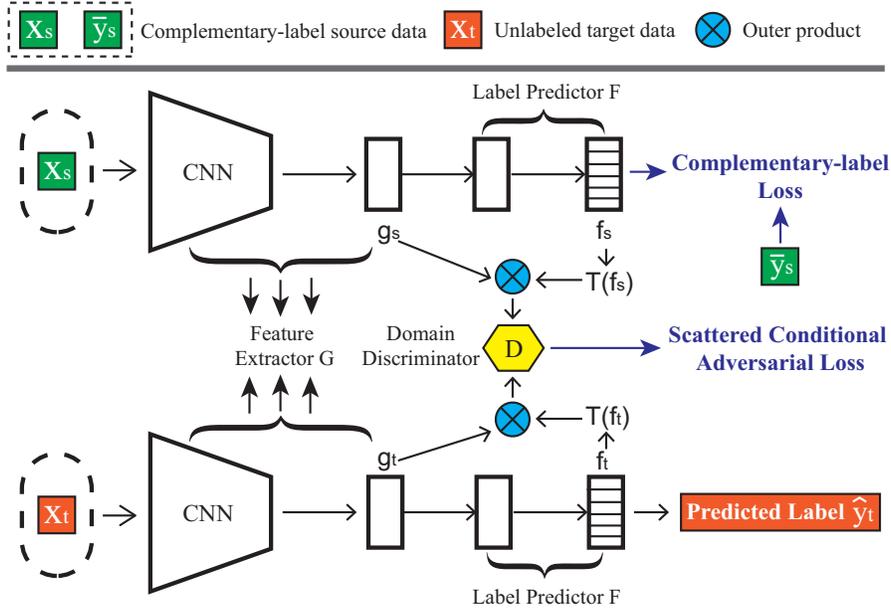}}
		\end{center}
		\vspace{-1em}
		\caption{{Overview of the proposed \emph{\underline{c}omplementary \underline{l}abel advers\underline{ari}al \underline{net}work} (CLARINET). It consists of feature extractor $G$, label predictor $F$ and conditional domain discriminator $D$. $g_s$ and $g_t$ are outputs of $G$, representing extracted features of source and target data. $f_s$ and $f_t$ represent classifier predictions. $T$ is a sharpening function which we propose to scatter the classifier predictions. In Algorithm~\ref{alg:algorithm}}, we show how to use two losses mentioned in this figure to train CLARINET.}
		\label{fig: CLARINET}
\end{figure*}

\section{Two-step Approach}

To solve the problem that existing UDA methods cannot be applied to complementary-label based UDA problems directly, a straightforward way is to apply a two-step strategy. Namely, we could  sequentially combine complementary-label learning methods and existing UDA methods. Algorithm \ref{alg:two} presents how we realize two-step approach for CC-UDA tasks specifically. In two-step approach, we first use the complementary-label learning algorithm to train a classifier on the complementary-label source data (line 1). Then, we take advantage of the classifier to assign pseudo labels for source domain data (line 2). Finally, we train the target-domain classifier with pseudo-label source data and unlabeled target data using existing UDA methods (line 3). In this way, we can transfer knowledge from the newly formed pseudo-label source data to unlabeled target data. As for PC-UDA tasks, we could combine the pseudo-label source data gotten following the first two steps and existing true-label source data together to train the target-domain classifier.

Nevertheless, the pseudo-label source data contains noise as complementary-label learning algorithms cannot be trained to produce a completely accurate classifier. As the noise will bring poor domain-adaptation performance \cite{liu2019butterfly}, the two-step approach is a suboptimal choice. To solve this problem, we consider implementing both complementary label learning and unsupervised domain adaptation in a network. In this way, the network will always try to classify source domain data accurately during the adaptation procedure. Besides, we consider using entropy condition to make the transfer process mainly based on the classification results with high confidence, which can largely eliminate the noise effect compared with the two-step approach.  Therefore, we propose a powerful one-step solution to complementary-label based UDA, CLARINET.

\section{CLARINET: Powerful One-step Approach}

The proposed CLARINET (as shown in Figure \ref{fig: CLARINET}) realizes distributional adaptation in an adversarial way, which mainly consists of feature extractor $G$, label predictor $F$ and domain discriminator $D$. By working adversarially to domain discriminator $D$, feature extractor $G$ encourages domain-invariant features to emerge. Label predictor $F$ are trained to discriminate different classes based on such features.

In this section, we first introduce two losses used to train CLARINET, complementary-label loss and scattered conditional adversarial loss. Then the whole training procedure of CLARINET is presented. Finally, we show how to adjust CLARINET for PC-UDA tasks if a small amount of true-label source data are available.

\subsection{Loss Function in CLARINET}
\label{sec:Loss}
In this subsection, we introduce how to compute the two losses mentioned above in CLARINET after obtaining mini-batch $\overline{d}_s$ from $\overline{D}_s$ and  $d_t$ from $D_t$.

\subsubsection{Complementary-label Loss.} 
% We follow the complementary-label classification risk of gradient ascent version in \cite{ishida2018complementary}. After draw a mini-batch $\check{D}$, 
It is designed to reduce the source classification error based on complementary-label data (the first part in the bound). We first divided $\overline{d}_s$ into $K$ disjoint subsets according to the complementary labels in $\overline{d}_s$,
\begin{equation}
    \overline{d}_s=\cup_{k=1}^K \overline{d}_{s,k},~\overline{d}_{s,k}=\{(\mathbf{x}_k^i,\mathbf{y}_k)\}_{i=1}^{\overline{n}_{s,k}},
\label{E1}
\end{equation}
where $\overline{d}_{s,k} \cap \overline{d}_{s,k'}=\varnothing$ if $k\neq{k'}$ and $\overline{n}_{s,k}=|\overline{d}_{s,k}|$. Then, following Eq. \eqref{main}, the complementary-label loss on $\overline{d}_{s,k}$ is

\begin{equation}
\begin{split}
    {\overline{L}}_{s}(G,F,\overline{d}_{s,k})=-&(K-1)\frac{{\overline\pi}_k}{\overline{n}_{s,k}}\sum_{i=1}^{\overline{n}_{s,k}}\ell(F\circ{G}(\mathbf{x}_k^i),\mathbf{y}_k) \\
    +&\sum_{j=1}^{K}\frac{{\overline\pi}_j}{\overline{n}_{s,j}}\sum_{l=1}^{\overline{n}_{s,j}}\ell(F\circ{G}(\mathbf{x}_j^l),\mathbf{y}_k),
\end{split}
\label{E2}
\end{equation}
where $\ell$ can be any loss and we use the cross-entropy loss, ${\overline\pi}_k$ is the proportion of the samples complementary-labeled $k$. The total complementary-label loss on $\overline{d}_s$ is as follows.
\begin{equation}
    {\overline{L}}_s(G,F,\overline{d}_s)=\sum_{k=1}^K{\overline{L}}_s(G,F,\overline{d}_{s,k}).
\label{E3}
\end{equation}

As shown in Section \ref{Theory}, the complementary-label loss (i.e., Eq.~\eqref{E3}) is an unbiased estimator of the true-label-data risk.  Namely, the minimizer of complementary-label loss agrees with the minimizer of the true-label-data risk with no constraints on the loss $\ell$ and model $F\circ{G}$ \cite{pmlr-v97-ishida19a}.

\begin{remark}
Due to the negative part in ${\overline{L}}_s(G,F,\overline{d}_s)$,  minimizing it directly will cause over-fitting \cite{kiryo2017positive}. To overcome this problem, we use a correctional way \cite{pmlr-v97-ishida19a} to minimize ${\overline{L}}_s(G,F,\overline{d}_s)$ (lines $7$-$13$ in Algorithm \ref{alg:algorithm}).
\end{remark}

\subsubsection{Scattered Conditional Adversarial Loss.} It is designed to reduce distribution discrepancy distance between two domains (the third part in the bound). Adversarial domain adaptation methods \cite{ganin2016domain,agresti2019} is inspired by Generative Adversarial Networks (GANs) \cite{goodfellow2014generative}. Normally, a domain discriminator is learned to distinguish the source domain and the target domain, while the label predictor learns transferable representations that are indistinguishable by the domain discriminator. Namely, the final classification decisions are made based on features that are both discriminative and invariant to the change of domains \cite{ganin2016domain}. It is an efficient way to reduce distribution discrepancy distance between the marginal distributions.

However, when data distributions have complex multimodal structures, which is a real scenario due to the nature of multi-class classification, adapting only the feature representation is a challenge for adversarial
networks. Namely, even the domain discriminator is confused, we could not confirm the two distributions are sufficiently
similar \cite{arora2017generalization}. 

According to \cite{song2009hilbert}, it is significant to capture multimodal structures of distributions using cross-covariance dependency between the features and classes (i.e., true labels). Since there are no true-label target data in UDA, CDAN adopts outer product of feature representations and classifier predictions (i.e., outputs of the softmax layer) as new features of two domains \cite{long2018conditional}, which is inspired by Conditional Generative Adversarial Networks (CGANs) \cite{mirza2014conditional,odena2017conditional}. The newly constructed features have shown great ability to discriminate source and target domains, since classifier predictions of true-label source data are dispersed, expressing the predicted goal clearly.

However, in the complementary-label classification mode, we observe that the predicted probability of each class (i.e., each element of $\bm{f_s}$ in Figure~\ref{fig: CLARINET})
is relatively close. Namely, it is hard to find significant predictive preference from the classifier predictions. According to \cite{song2009hilbert}, this kind of predictions cannot provide sufficient information to capture the multimodal structure of distributions. To fix it, we add a sharpening function $T$ to scatter the predicted probability (the output of ${\bm f}=[f_1,...,f_K]^T$ after Softmax function,  $\bm f$ could be $\bm{f_s}$ or $\bm{f_t}$ in Figure~\ref{fig: CLARINET}).

In \cite{goodfellow2016deep}, a common approach of adjusting the “temperature” of this categorical distribution is defined as follows,

\begin{equation}
    T({\bm f})=\left [\frac{{f}_1^{\frac{1}{l}}}{\sum_{j=1}^{K}{f}_j^{\frac{1}{l}}},...,\frac{{f}_k^{\frac{1}{l}}}{\sum_{j=1}^{K}{f}_j^{\frac{1}{l}}},...,\frac{{f}_K^{\frac{1}{l}}}{\sum_{j=1}^{K}{f}_j^{\frac{1}{l}}}\right]^T.
\label{eq:T1}
\end{equation}
%\begin{equation}
%    T_1(P,l)_i=\frac{P_i^{\f%rac{1}{l}}}{\sum_{j=1}^{K}P_j^{\frac{1}{l}}}.
%\label{eq:T1}
%\end{equation}
As $l \rightarrow 0$, the output of $T({\bm{f}})$ will approach a Dirac (“one-hot”) distribution \cite{berthelot2019mixmatch}. 
%Actually, we explored several ways to realize the sharpening function $T$ and we will introduce them in the section \ref{Sec:T}. Without special instructions, the sharpening function $T$ in the CLARINET refers to Eq.~\eqref{eq:T1}).   

\iffalse
To fix it, we propose a sharpening function $T$ to scatter the classifier predictions ${\bm f}=[f_1,...,f_K]^T$ ($\bm f$ could be $\bm{f_s}$ or $\bm{f_t}$ in Figure~\ref{fig: CLARINET}),
\begin{equation}
    T({ f}_k)=\textnormal{exp}\left(-\Big(\frac{\epsilon}{{ f}_k+{ \eta}}\Big)^r\right),~~~\forall k=1,...K,
\end{equation}
where ${\bm f}=F\circ G$, $r$ is an integer greater than $1$, ${ \eta}$ is a small number in case $f_k=0$ and $\epsilon$ is a threshold value. If $f_k$ is larger than $\epsilon$, the sharpening function $T$ will map $f_k$ to get closer to $1$, otherwise it will map $f_k$ to get closer to $0$.
\fi

Then to prioritize the discriminator on those easy-to-transfer examples, following \cite{long2018conditional}, we measure the uncertainty of the prediction for sample $\mathbf{x}$ by
\begin{equation}
    H(G,F,\mathbf{x})=-\sum_{k=1}^K{T(f_k(\mathbf{x}))}{\log{T(f_k(\mathbf{x}))}}.
\label{EH}
\end{equation}
The small result implies that $T(f_k(\mathbf{x}))$ is close to 0 or 1, which could be regarded as the prediction is with high confidence due to the existing of the  final softmax layer \cite{zhu2003semi}.

Thus the scattered conditional adversarial loss is as follows,
\begin{equation}
%\setlength{\abovedisplayskip}{2pt}
%\setlength{\belowdisplayskip}{3pt}
%\resizebox{1\linewidth}{!}{$
\begin{split}
    \displaystyle
    L_{adv}(G,F,D,\overline{d}_s,d_t)=&\frac{\sum_{\mathbf{x}\in \overline{d}_{s}[X]}\omega_{\overline{s}}(\mathbf{x})\log(D({\bm g(\mathbf{x})}))}{\sum_{\mathbf{x}\in \overline{d}_{s}[X]}\omega_{\overline{s}}(\mathbf{x})}
    \\+&\frac{\sum_{\mathbf{x}\in d_t}\omega_t(\mathbf{x}){\log(1-D({\bm g(\mathbf{x})}))}}{\sum_{\mathbf{x}\in d_t}\omega_t(\mathbf{x})},
\end{split}
%$}
\label{adv1}
\end{equation}
where $\omega_{\overline{s}}(\mathbf{x})$ and $\omega_{t}(\mathbf{x})$ are $1+e^{-H(G,F,\mathbf{x})}$, ${\bm g}(\mathbf{x})$ is $G(\mathbf{x})\otimes{T(F\circ{G}(\mathbf{x})})$ and $\overline{d}_{s}[X]$ is the feature part of $\overline{d}_s$.

\begin{algorithm}[tb]
\small
\caption{CLARINET for CC-UDA Tasks}
\label{alg:algorithm}
\textbf{Input}: $\overline{D}_s=\{(\mathbf{x}_s^i,\overline{\mathbf{y}}_s^i)\}^{\overline{n}_s}_{i=1}$, $D_t=\{\mathbf{x}_t^i\}^{n_t}_{i=1}$.\\
\textbf{Parameters}: learning rate $\gamma_1$ and $\gamma_2$, epoch $T_{max}$, start epoch $T_s$, iteration $N_{max}$, class number $K$, tradeoff $\lambda$, network parameter $\theta_{F\circ{G}}$ and $\theta_D$.\\
\textbf{Output}: the neural network $F\circ{G}$, namely the target domain classifier for $D_t$.
\begin{algorithmic}[1] %[1] enables line numbers
\STATE \textbf{Initialize} $\theta_{F\circ{G}}$ and $\theta_D$;
\FOR{$t=1,2……T_{max}$}
\STATE \textbf{Shuffle} the training set $\overline{D}_s$, ${D}_t$;
\FOR{$N=1,2……N_{max}$}
\STATE \textbf{Fetch} mini-batch $\overline{d}_s$, $d_t$ from $\overline{D}_s$, ${D}_t$;
\STATE \textbf{Divide} $\overline{d}_s$ into $\{\overline{d}_{s,k}\}_{k=1}^K$;
\STATE \textbf{Calculate} $\{\overline{L}_s(G,F,\overline{d}_{s,k})\}_{k=1}^K$ using Eq.~\eqref{E2}, and $\overline{L}_s(G,F,\overline{d}_{s})$ using Eq.~\eqref{E3};
\IF{$\min_k\{\overline{L}_s(G,F,\overline{d}_{s,k})\}_{k=1}^{K}\geq 0$}
\STATE \textbf{Update} $\theta_{F\circ{G}}=\theta_{F\circ{G}}-\gamma_1\triangledown \overline{L}_s(G,F,\overline{d}_{s})$;
\ELSE
\STATE \textbf{Calculate} $\overline{L}_{neg}=\sum_{k=1}^K\min\{0,\overline{L}_s(G,F,\overline{d}_{s,k})\}$;
\STATE \textbf{Update} $\theta_{F\circ{G}}=\theta_{F\circ{G}}+\gamma_1\triangledown \overline{L}_{neg}$;
\ENDIF
% \ENDFOR
\IF{$t > T_s$}
\STATE \textbf{Calculate} $L_{adv}(G,F,D,\overline{d}_s,d_t)$ using Eq.~\eqref{adv1};
\STATE \textbf{Update} $\theta_D=\theta_D-\gamma_2\triangledown{L_{adv}(G,F,D,\overline{d}_s,d_t)}$;
\STATE \textbf{Update} $\theta_{F\circ{G}}=\theta_{F\circ{G}}+\gamma_2\lambda\triangledown{L_{adv}(G,F,D,\overline{d}_s,d_t)}$;
\ENDIF
\ENDFOR
\ENDFOR
\end{algorithmic}
\end{algorithm}

\subsection{Training Procedure of CLARINET}

Based on two losses proposed in Section~\ref{sec:Loss}, in CLARINET, we try to solve the following optimization problem,
\begin{equation}
\label{eq:minmax}
\begin{split}
    \min_{G,F}&~\overline{L}_s(G,F,\overline{D}_s)-\lambda L_{adv}(G,F,D,\overline{D}_s,D_t),\\
    \min_D&~ L_{adv}(G,F,D,\overline{D}_s,D_t),
\end{split}
\end{equation}
where $D$ tries to distinguish the samples from different domains by minimizing $L_{adv}$, while $F\circ{G}$ wants to maximize the $L_{adv}$ to make domains indistinguishable.   
To solve the minimax optimization problem in Eq.~\eqref{eq:minmax}, we add a gradient reversal layer \cite{ganin2016domain} between the domain discriminator and the classifier, which multiplies the gradient by a negative constant (-$\lambda$) during the back-propagation. $\lambda$ is a hyper-parameter between the two losses to tradeoff source risk and domain discrepancy. 

The training procedures of CLARINET are shown in Algorithm~\ref{alg:algorithm}. First, we initialize the whole network (line $1$) and shuffle the training set (line $3$). During each epoch, after minbatch $\overline{d}_s$ and $d_t$ are fetched (line $5$), we divide the source mini-batch $\overline{d}_s$ into $\{\overline{d}_{s,k}\}_{k=1}^K$ using Eq.~\eqref{E1} (line $6$). Then, $\{\overline{d}_{s,k}\}_{k=1}^K$ are used to calculate the complementary-label loss for each class (i.e., $\{\overline{L}_s(G,F,\overline{d}_{s,k})\}_{k=1}^{K}$) and the whole complementary-label loss $\overline{L}_s(G,F,\overline{d}_s)$ (line $7$). 

If $\min_k\{\overline{L}_s(G,F,\overline{d}_{s,k})\}_{k=1}^{K}\geq 0$, we calculate the gradient $\triangledown \overline{L}_s(G,F,\overline{d}_s)$ and update parameters of $G$ and $F$ using gradient descent (lines $8$-$9$). 
Otherwise, we sum negative elements in $\{\overline{L}_s(G,F,\overline{d}_{s,k})\}_{k=1}^{K}$ as $\overline{L}_{neg}$ (line $11$) and calculate the gradient with $\triangledown\overline{L}_{neg}$ (line $12$). 
Then, we update parameters of $G$ and $F$ using gradient ascent (line $12$), which is suggested by \cite{pmlr-v97-ishida19a}. 
When the number of epochs (i.e., $t$) is over $T_s$, we start to update parameters of $D$ (line $14$). We calculate the scattered conditional adversarial loss $L_{adv}$ (line $15$). Then, $L_{adv}$ is minimized over $D$ (line $16$), but maximized over $F\circ{G}$ (line $17$) for adversarial training.

\subsection{CLARINET for PC-UDA Tasks}

For PC-UDA tasks, we have both complementary-label data and true-label data in the source domain. In such cases, we want to leverage both kinds of labeled source data to help realize better adaptation results. The two loss functions mentioned in Section \ref{sec:Loss} are adjusted as follows.

After obtaining mini-batch $d_s$ from $D_s$, we could calculate the classification loss based on true-label data by
\begin{equation}
    L_s(G,F,d_s)=\ell(F\circ{G(\mathbf{x}_i),\mathbf{y}_i}),
\label{eq:Lt}
\end{equation}
where $\ell$ is cross-entropy loss, $d_s=\{(\mathbf{x}_i,y_i)\}_{i=1}^{n_s'}$ and $n_s'=|d_s|$.
We could use a convex combination of classification risks derived from true-label data and complementary-label data to replace the oral complementary-label based only classification risk shown as following.
\begin{equation}
    L_c=\alpha L_s(G,F,d_s)+(1-\alpha)\overline{L}_s(G,F,\overline{d}_s),
\end{equation}
where $\alpha$ depends on the cost of labeling the two kind of data. 

The new scattered conditional adversarial loss for PC-UDA tasks is as follows.
\begin{equation}
\begin{split}
    \displaystyle
    &L_{adv}(G,F,D,d_s,\overline{d}_s,d_t)\\
    =~&    \frac{\sum_{\mathbf{x}\in {d}_{s}[X]}\omega_{{s}}(\mathbf{x})\log(D({\bm g(\mathbf{x})}))}{\sum_{\mathbf{x}\in {d}_{s}[X]}\omega_{{s}}(\mathbf{x})}
    \\+~&    \frac{\sum_{\mathbf{x}\in \overline{d}_{s}[X]}\omega_{\overline{s}}(\mathbf{x})\log(D({\bm g(\mathbf{x})}))}{\sum_{\mathbf{x}\in \overline{d}_{s}[X]}\omega_{\overline{s}}(\mathbf{x})}
    \\+~&    \frac{\sum_{\mathbf{x}\in d_t}\omega_t(\mathbf{x}){\log(1-D({\bm g(\mathbf{x})}))}}{\sum_{\mathbf{x}\in d_t}\omega_t(\mathbf{x})},
\end{split}
\label{eq:advP}
\end{equation}
%where $\omega_s=1+e^{-H(G,F,{d}_s)}$, $\omega_{\overline{s}}=1+e^{-H(G,F,\overline{d}_s)}$ and $\omega_t=1+e^{-H(G,F,d_t)}$ calculated based on Eq.~\eqref{EH}. 
where $\omega_{{s}}(\mathbf{x})$, $\omega_{\overline{s}}(\mathbf{x})$ and $\omega_{t}(\mathbf{x})$ are $1+e^{-H(G,F,\mathbf{x})}$, ${\bm g}(\mathbf{x})$ is $G(\mathbf{x})\otimes{T(F\circ{G}(\mathbf{x})})$, ${d}_{s}[X]$ and $\overline{d}_{s}[X]$ is the feature part of ${d}_s$ and $\overline{d}_s$. The entire training procedures of CLARINET for PC-UDA are shown in Algorithm~\ref{alg:PC-UDA}.

\begin{algorithm}[tb]
\small
\caption{CLARINET for PC-UDA Tasks}
\label{alg:PC-UDA}
\textbf{Input}: ${D}_s=\{(\mathbf{x}_s^i,\mathbf{y}_s^i)\}_{i=1}^{n_s}$, $\overline{D}_s=\{(\mathbf{x}_s^i,\overline{\mathbf{y}}_s^i)\}^{\overline{n}_s}_{i=1}$, $D_t=\{\mathbf{x}_t^i\}^{n_t}_{i=1}$.\\
\textbf{Parameters}: learning rate $\gamma_1$ and $\gamma_2$, epoch $T_{max}$, start epoch $T_s$, iteration $N_{max}$, class number $K$, tradeoff $\lambda$ and $\alpha$,  network parameter $\theta_{F\circ{G}}$ and $\theta_D$.\\
\textbf{Output}: the neural network $F\circ{G}$, namely the target domain classifier for $D_t$.
%\textbf{Output}: the neural network $G\circ{F}$, the target domain classifier for $D_t$.
\begin{algorithmic}[1] %[1] enables line numbers
\STATE \textbf{Initialize} $\theta_{F\circ{G}}$ and $\theta_D$;
\FOR{$T=1,2……T_{max}$}
\STATE \textbf{Shuffle} the training set ${D}_s$, $\overline{D}_s$, ${D}_t$;
\FOR{$N=1,2……N_{max}$}
\STATE \textbf{Fetch} minibatch $d_s$, $\overline{d}_s$, $d_t$ from $D_s$, $\overline{D}_s$, ${D}_t$;
% \STATE \textbf{Obtain} $\bm{f_s}=F(\check D_s)$, $\bm{f_t}=F(\check D_t)$, 
% \STATE \textbf{Obtain} $\bm{g_s}=F\circ{G}(\check D_s)$,  $\bm{g_t}=F\circ{G}(\check D_t)$.
\STATE \textbf{Calculate} $L_s(G,F,d_s)$ using Eq.~\eqref{eq:Lt};
\STATE \textbf{Update} $\theta_{F\circ{G}}=\theta_{F\circ{G}}-\gamma_1\alpha\triangledown L_s(G,F,d_s)$;
\STATE \textbf{Divide} $\overline{d}_s$ into $\{\overline{d}_{s,k}\}_{k=1}^K$;
\STATE \textbf{Calculate} $\{\overline{L}_s(G,F,\overline{d}_{s,k})\}_{k=1}^K$ using Eq.~\eqref{E2}, and $\overline{L}_s(G,F,\overline{d}_{s})$ using Eq.~\eqref{E3};
% , and $\widetilde{L}_{com}^b(G,\check{D}_{s})$ using Eq.~\eqref{};
% \FOR{$k=1,\dots, K$}
\IF{$\min_k\{\overline{L}_s(G,F,\overline{d}_{s,k})\}_{k=1}^{K}\geq 0$}
\STATE \textbf{Update} $\theta_{F\circ{G}}=\theta_{F\circ{G}}-\gamma_1(1-\alpha)\triangledown \overline{L}_s(G,F,\overline{d}_{s})$;
\ELSE
\STATE \textbf{Calculate} $\overline{L}_{neg}=\sum_{k=1}^K\min\{0,\overline{L}_s(G,F,\overline{d}_{s,k})\}$;
\STATE \textbf{Update} $\theta_{F\circ{G}}=\theta_{F\circ{G}}+\gamma_1(1-\alpha)\triangledown \overline{L}_{neg}$;
\ENDIF
% \ENDFOR
\IF{$T > T_s$}
% \STATE \textbf{Calculate} $F(\check D_s)\otimes{T(F\circ{G}(\check D_s))}$, $\omega_s$.
% \STATE \textbf{Calculate}
% $F(\check D_t)\otimes{T(F\circ{G}(\check D_t))}$, $\omega_t$. 
\STATE \textbf{Calculate} $L_{adv}(G,F,D,d_s,\overline{d}_s,d_t)$ using Eq.~\eqref{eq:advP};
\STATE \textbf{Update} $\theta_D=\theta_D-\gamma_2\triangledown{L_{adv}(G,F,D,d_s,\overline{d}_s,d_t)}$;
\STATE \textbf{Update} $\theta_{F\circ{G}}+=\gamma_2\lambda\triangledown{L_{adv}(G,F,D,d_s,\overline{d}_s,d_t)}$;
\ENDIF
\ENDFOR
\ENDFOR
\end{algorithmic}
\end{algorithm}

\iffalse
In practical situations, we may have some true-label source data. In such cases, we can leverage both kinds of data to obtain a more accurate classifier by using a convex combination of the classification risks derived from true-label source data and complementary-label source data. 
\begin{equation}
    L_c=\alpha L_s(G,F,d_s)+(1-\alpha)\overline{L}_s(G,F,\overline{d}_s),
\end{equation}
where $d_s$ is a mini-batch drawn from true-label source data, $\alpha$ is a hyper-parameter that interpolates between the two risks. In our experiments (Section~\ref{true ex}), we set $\alpha$ as follows, 
\begin{equation}
    \alpha = {n_s}\Big/{\Big(n_s+\frac{\overline{n}_s}{K-1}\Big)},
\end{equation}
where $n_s$ and $\overline{n}_s$ are the number of true-label source data and complementary-label source data, respectively.
\fi
\section{Experiments}

In this section, we conducted extensive evaluations of the proposed CLARINET on several common transfer tasks against many varients of state-of-the-art transfer learning methods (e.g. two-step approach). 

\subsection{Datasets and Tasks}

We investigate seven image and digits datasets: {\emph{CIFAR}} \cite{CIFAR}, {\emph{STL}} \cite{STL}, {\emph{MNIST}} \cite{lecun1998gradient}, {\emph{USPS}} \cite{DBLP:books/sp/HastieFT01}, {\emph{SVHN}} \cite{svhn}, {\emph{MNIST-M}}  \cite{DBLP:conf/icml/GaninL15} and {\emph{SYN-DIGITS}} \cite{DBLP:conf/icml/GaninL15}. We adopt the evaluation protocol of DANN \cite{ganin2016domain}, CDAN \cite{long2018conditional}), ATDA \cite{saito2017asymmetric}, and DIRT-T \cite{DBLP:conf/iclr/ShuBNE18} with seven transfer tasks: {\emph{CIFAR}} to {\emph{STL}} ($\emph{C}$$\rightarrow$ $\emph{T}$), {\emph{MNIST}} to {\emph{USPS}} ($\emph{M}$ $\rightarrow$ $\emph{U}$), {\emph{USPS}} to {\emph{MNIST}} ($\emph{U}$ $\rightarrow$ $\emph{M}$), {\emph{SVHN}} to {\emph{MNIST}} ($\emph{S}$ $\rightarrow$ $\emph{M}$),  {\emph{MNIST}} to {\emph{MNIST-M}} ($\emph{M}$ $\rightarrow$ $\emph{m}$), {\emph{SYN-DIGITS}} to {\emph{MNIST}} ($\emph{Y}$$\rightarrow$ $\emph{M}$) and {\emph{SYN-DIGITS}} to {\emph{SVHN}} ($\emph{Y}$$\rightarrow$ $\emph{S}$). 

We train our model using the training sets: {\emph{CIFAR}} ($45,000$), {\emph{STL}} ($4,500$), {\emph{MNIST}} ($60,000$), {\emph{USPS}} ($7,438$), {\emph{SVHN}} ($73,257$), {\emph{MNIST-M}} ($59,001$), {\emph{SYN-DIGITS}} ($479,400$). Evaluation is reported on the standard test sets: {\emph{STL}} ($7,200$), {\emph{MNIST}} ($10,000$), {\emph{USPS}} ($1,860$), {\emph{MNIST-M}} ($9,001$), {\emph{SVHN}} ($26,032$) (the numbers of images are in parentheses). %Note that, we generate complementary-label data according to \cite{pmlr-v97-ishida19a}.

Since all datasets carry true labels, following \cite{pmlr-v97-ishida19a}, we generate completely and partly complementary-label data. Generating complementary-label data is straightforward when the dataset is ordinary-labeled, as it reduces to just choosing a class randomly other than true class.
%Based on five commonly used datasets: {\emph{MNIST}} (\emph{M}), {\emph{USPS}} (\emph{U}), {\emph{SVHN}} (\emph{S}), {\emph{MNIST-M}} (\emph{m}) and {\emph{SYN-DIGITS}} (\emph{Y}), we verify efficacy of CLARINET on $6$ {BFUDA tasks}: $\emph{M}$ $\rightarrow$ $\emph{U}$, $\emph{U}$ $\rightarrow$ $\emph{M}$, $\emph{S}$ $\rightarrow$ $\emph{M}$,  $\emph{M}$ $\rightarrow$ $\emph{m}$, $\emph{Y}$$\rightarrow$ $\emph{M}$ and $\emph{Y}$$\rightarrow$ $\emph{S}$. Note that, we generate complementary-label data according to \cite{pmlr-v97-ishida19a}.

\subsection{Baselines}

We compare CLARINET with the following baselines: {\em gradient ascent complementary label learning} (GAC) \cite{pmlr-v97-ishida19a}, namely non-transfer method, and several two-step methods, which sequentially combine GAC with UDA methods (including DAN \cite{long2015learning}, DANN \cite{ganin2016domain} and CDAN \cite{long2018conditional}). Thus, we have four possible baselines: GAC, GAC+DAN, GAC+DANN and GAC+CDAN. For two-step methods, they share the same pseudo-label source data on each task. Note that, in this paper, we use the entropy conditioning variant of CDAN (CDAN\_E). 

%In two-step approach, GAC method is first used to assign pseudo labels for complementary-label source data. Then, we train the classifier with pseudo-label source data and unlabeled target data using UDA methods. Thus, we have four possible baselines: GAC, GAC+DAN (G+DAN), GAC+DANN (G+DANN) and GAC+CDAN (G+CDAN). For two-step methods, they share the same pseudo-label source data on each task. Note that, in this paper, we use the entropy conditioning variant of CDAN.  

\begin{figure*}[t]
\centering
\subfigure[CIFAR $\rightarrow$ STL.]{
\includegraphics[width=0.32\textwidth]{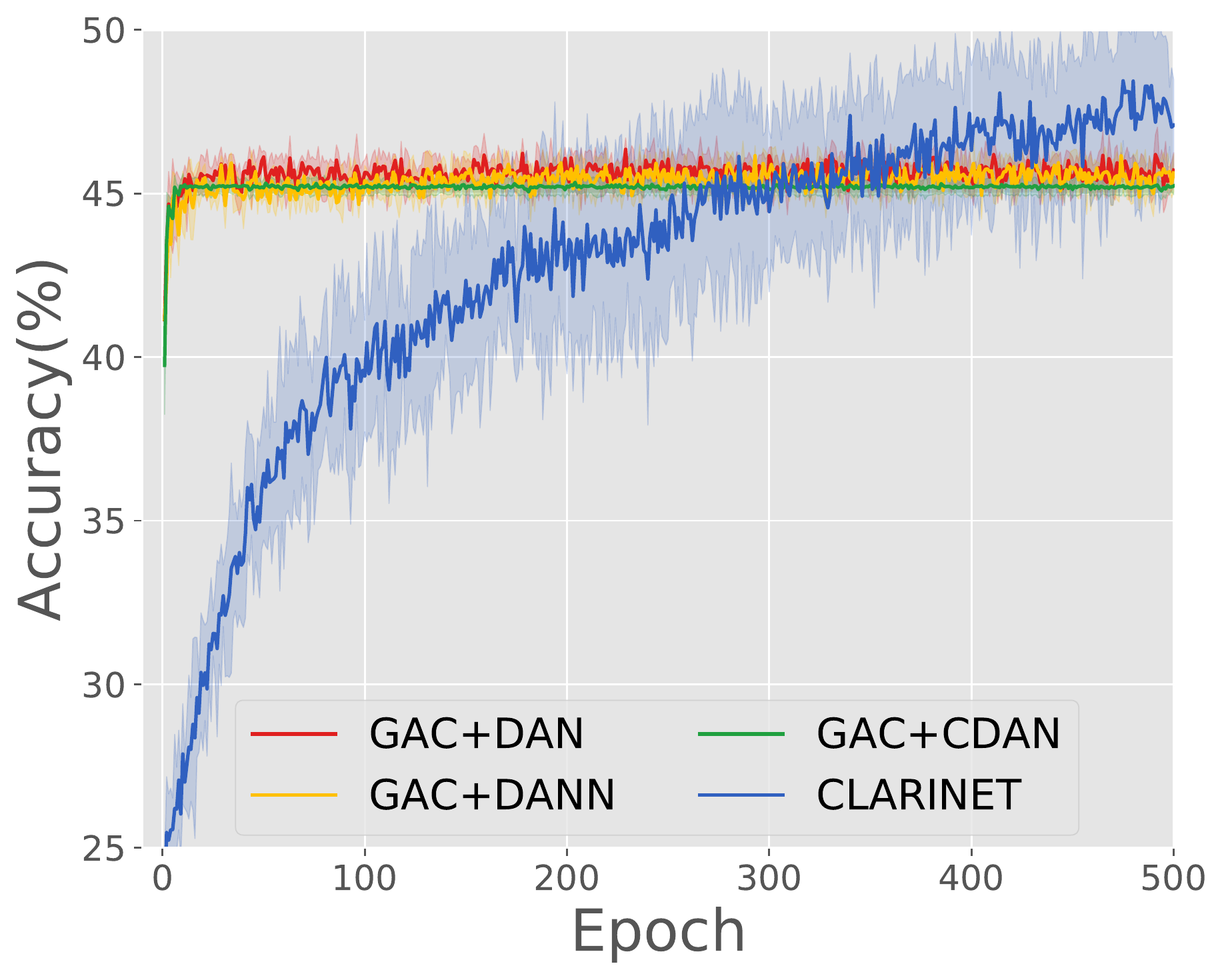}
%\caption{fig1}
}
%\quad
\subfigure[USPS $\rightarrow$ MNIST.]{
\includegraphics[width=0.3\textwidth]{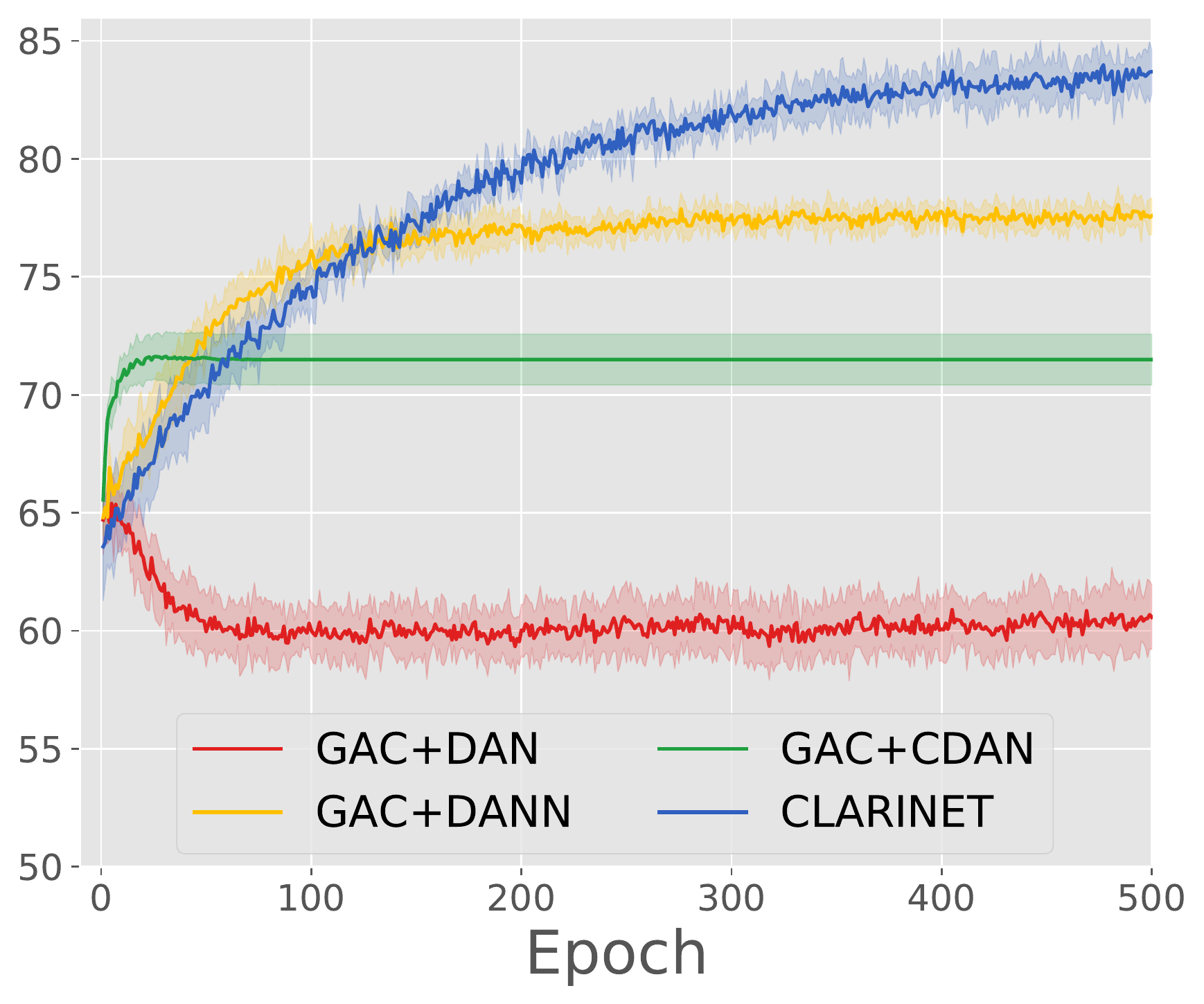}
}
\subfigure[MNIST $\rightarrow$ USPS.]{
\includegraphics[width=0.3\textwidth]{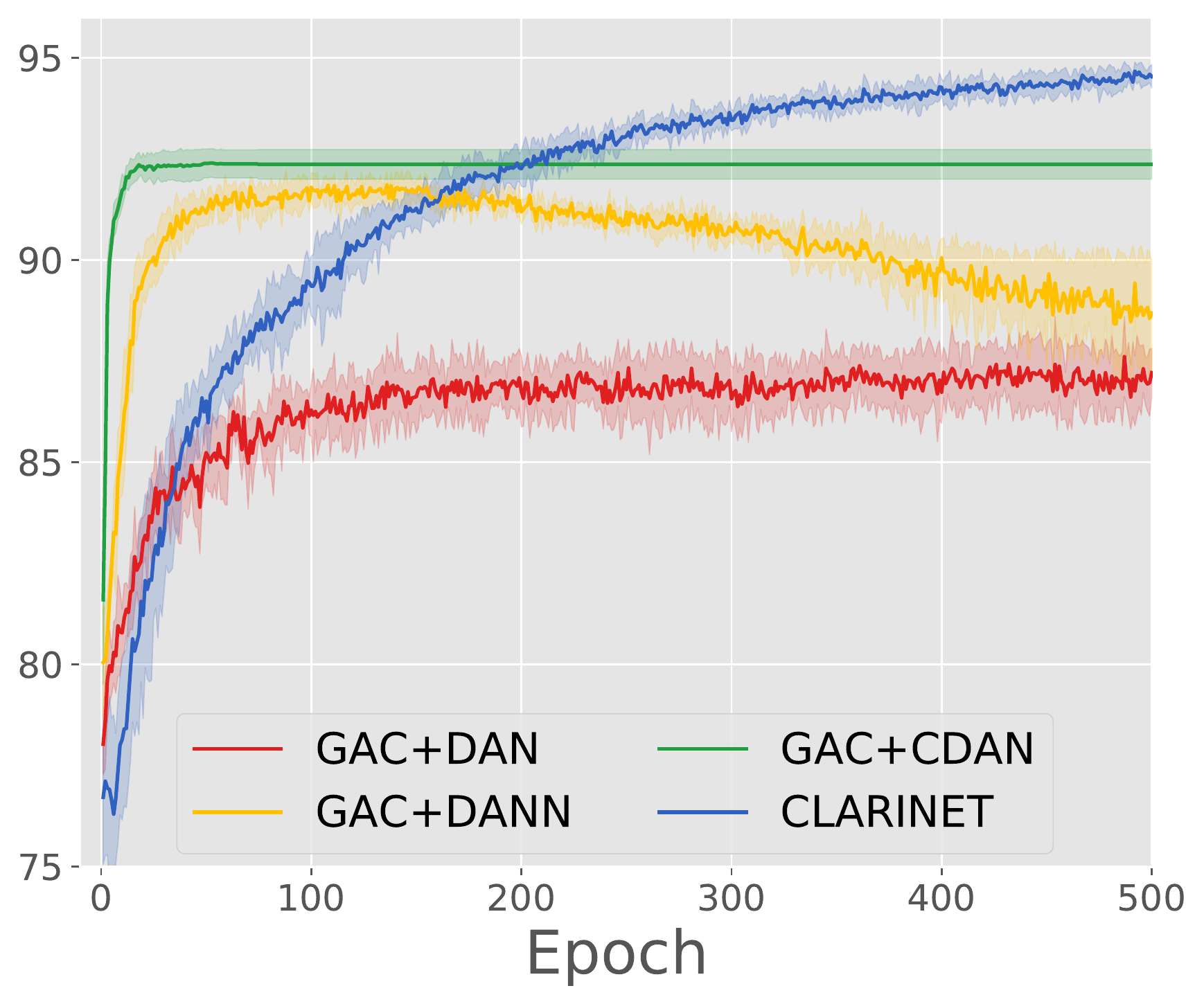}
}
\subfigure[SVHN $\rightarrow$ MNIST.]{
\includegraphics[width=0.32\textwidth]{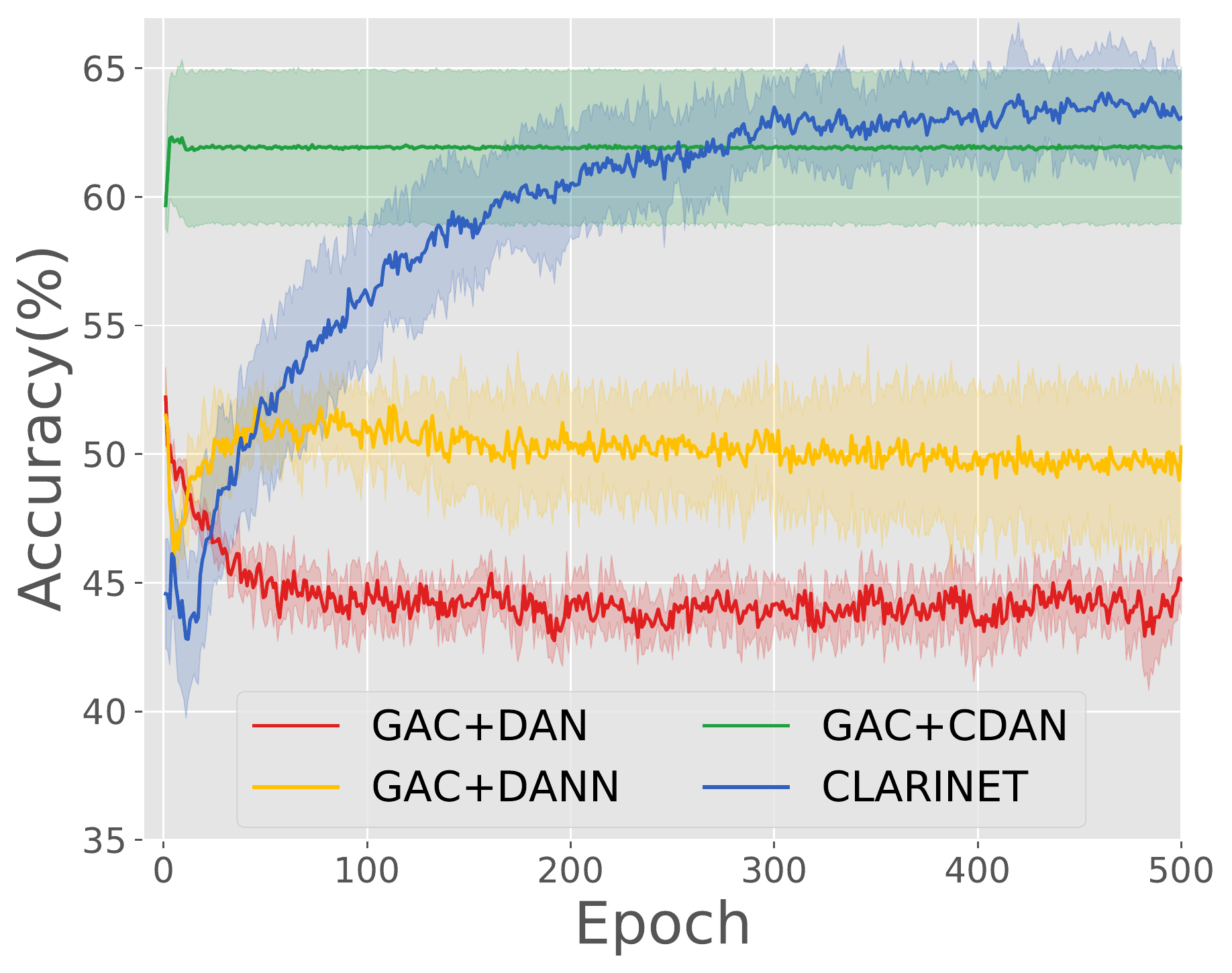}
%\caption{fig1}
}
%\quad
\subfigure[MNIST $\rightarrow$ MNIST-M.]{
\includegraphics[width=0.3\textwidth]{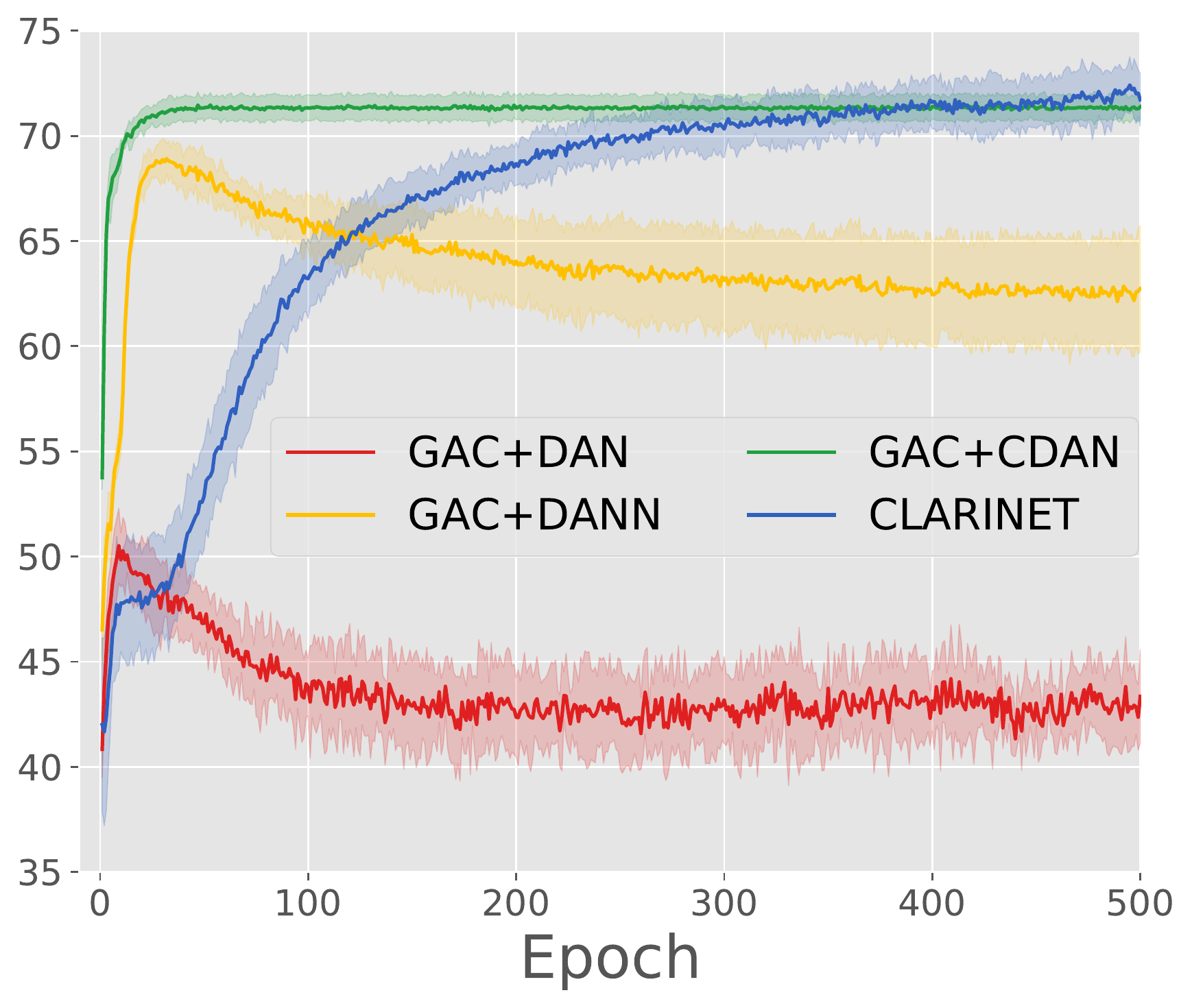}
}
\subfigure[SYND $\rightarrow$ MNIST.]{
\includegraphics[width=0.3\textwidth]{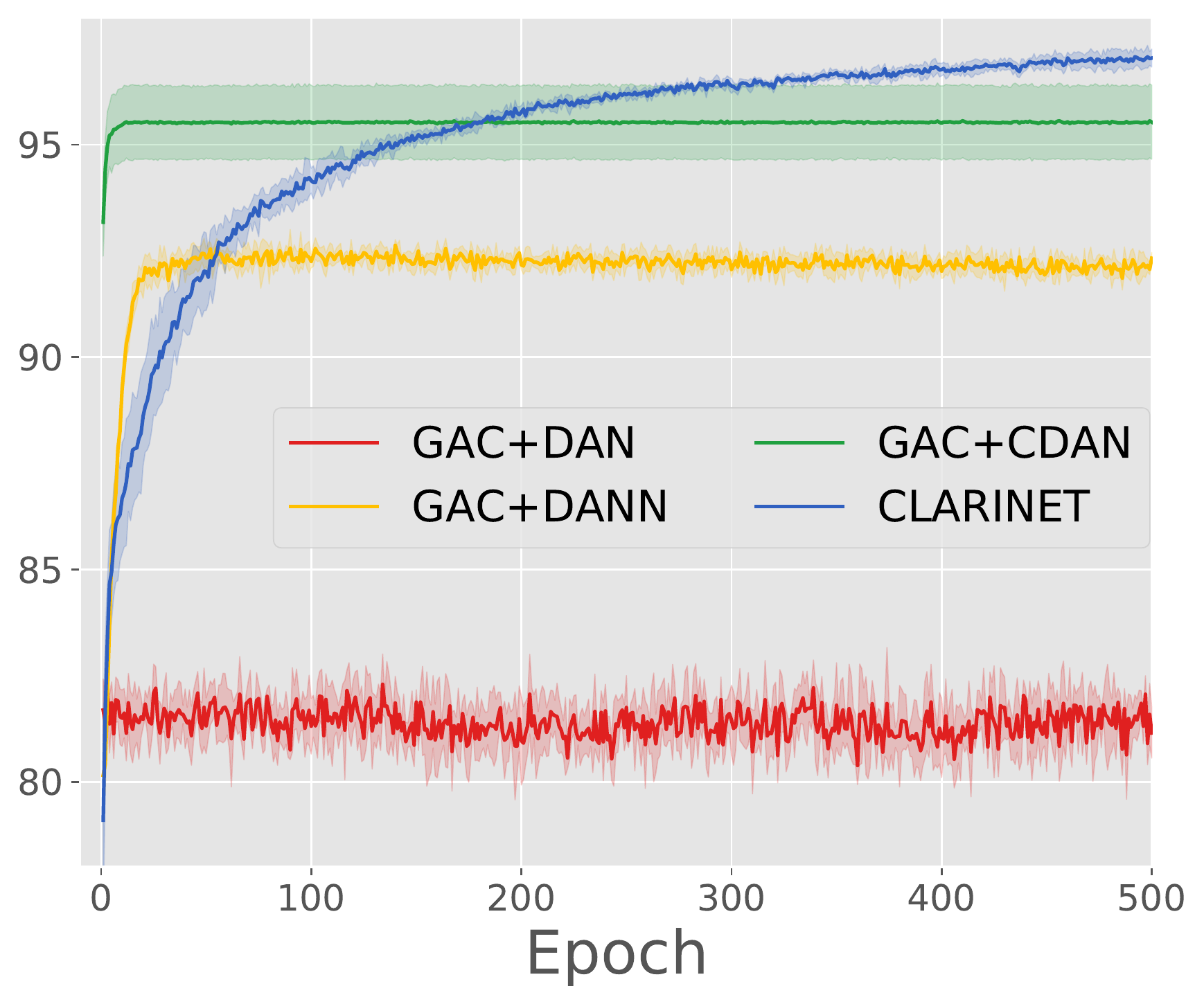}
}
%\quad
\subfigure[SYND $\rightarrow$ SVHN.]{
\includegraphics[width=0.32\textwidth]{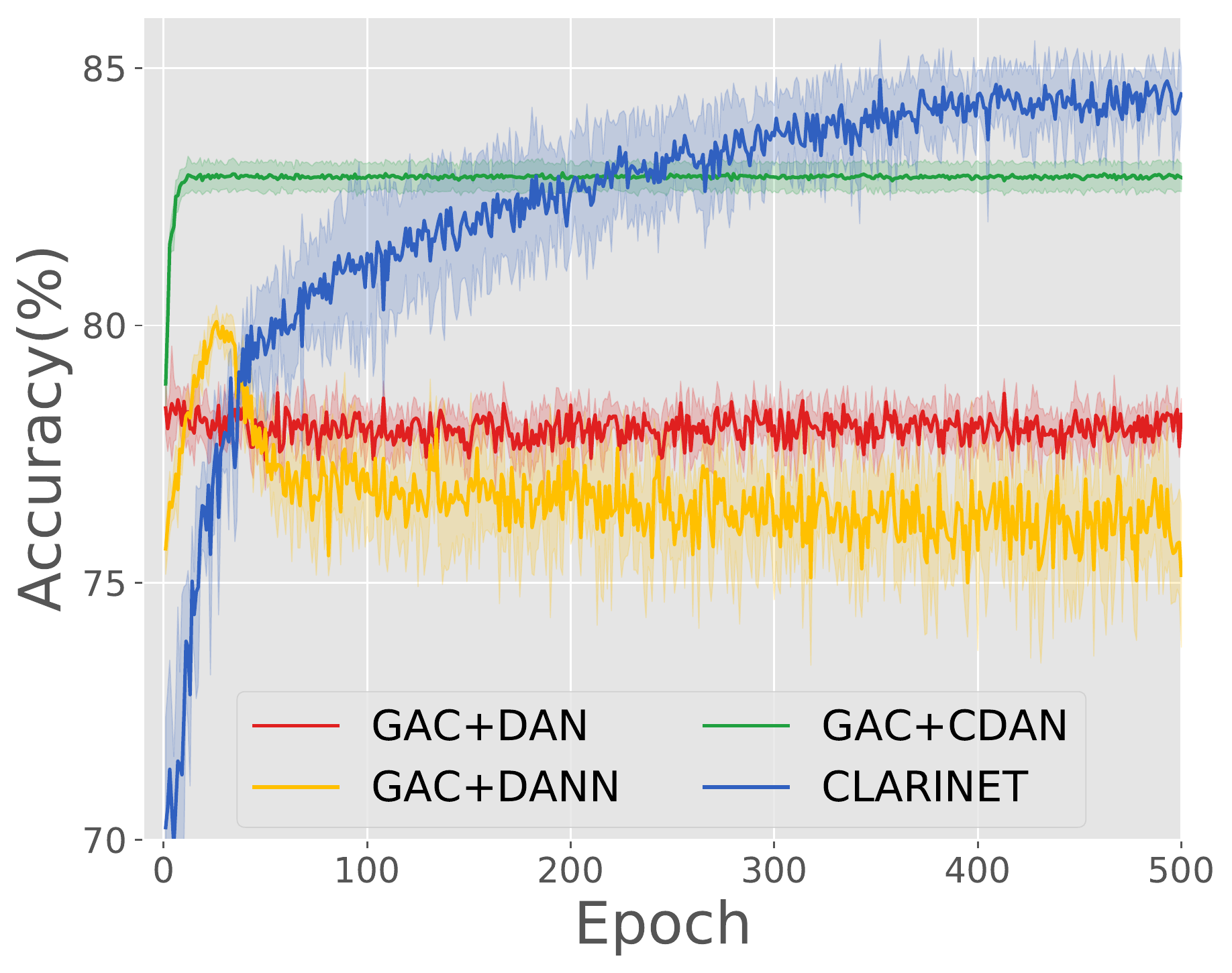}
}
% \quad
% \vspace{-1em}
\subfigure[True Label \emph{vs.} Complementary Label (SVHN $\rightarrow$ MNIST).]
{
\includegraphics[width=0.3\textwidth]{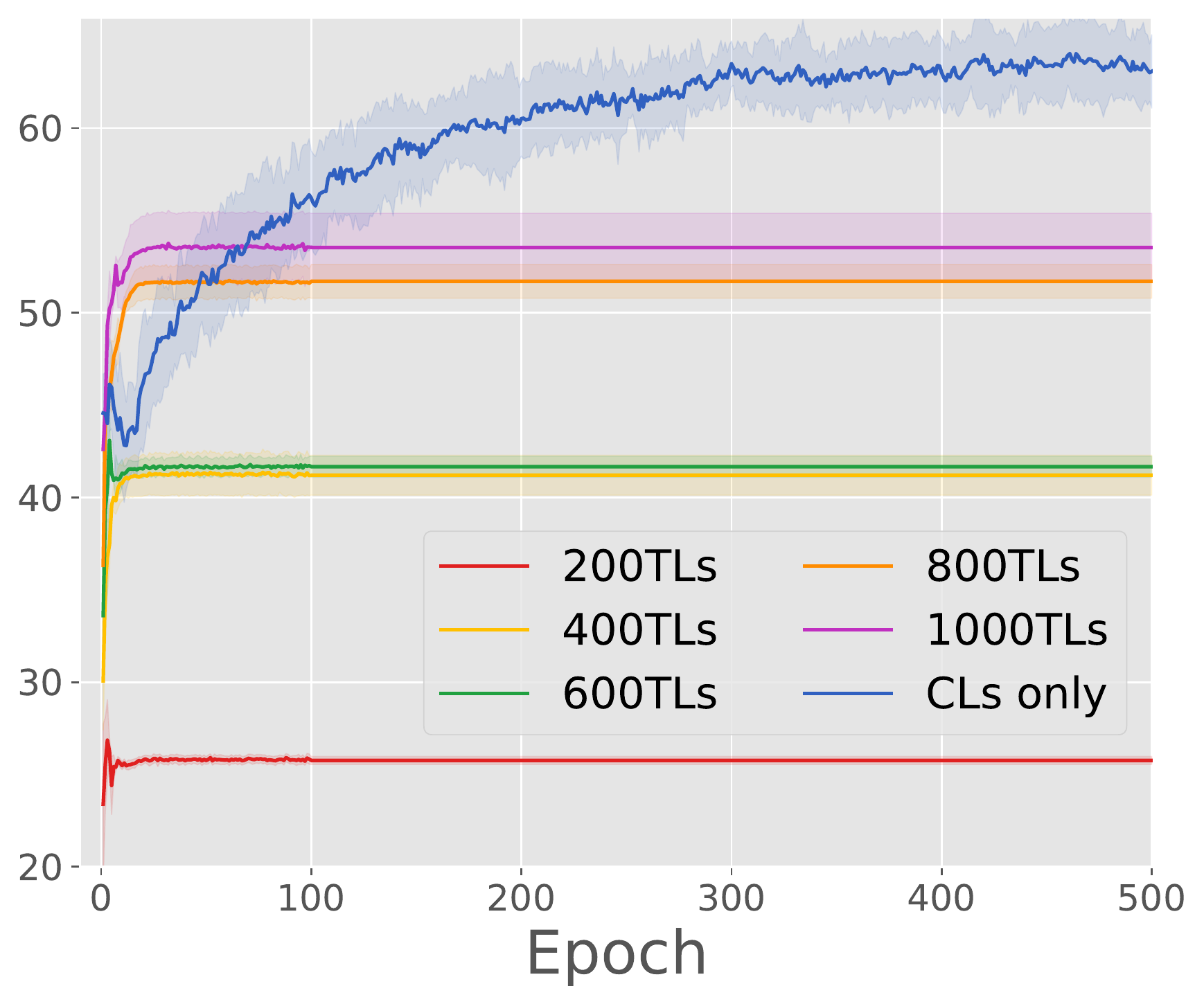}
\includegraphics[width=0.3\textwidth]{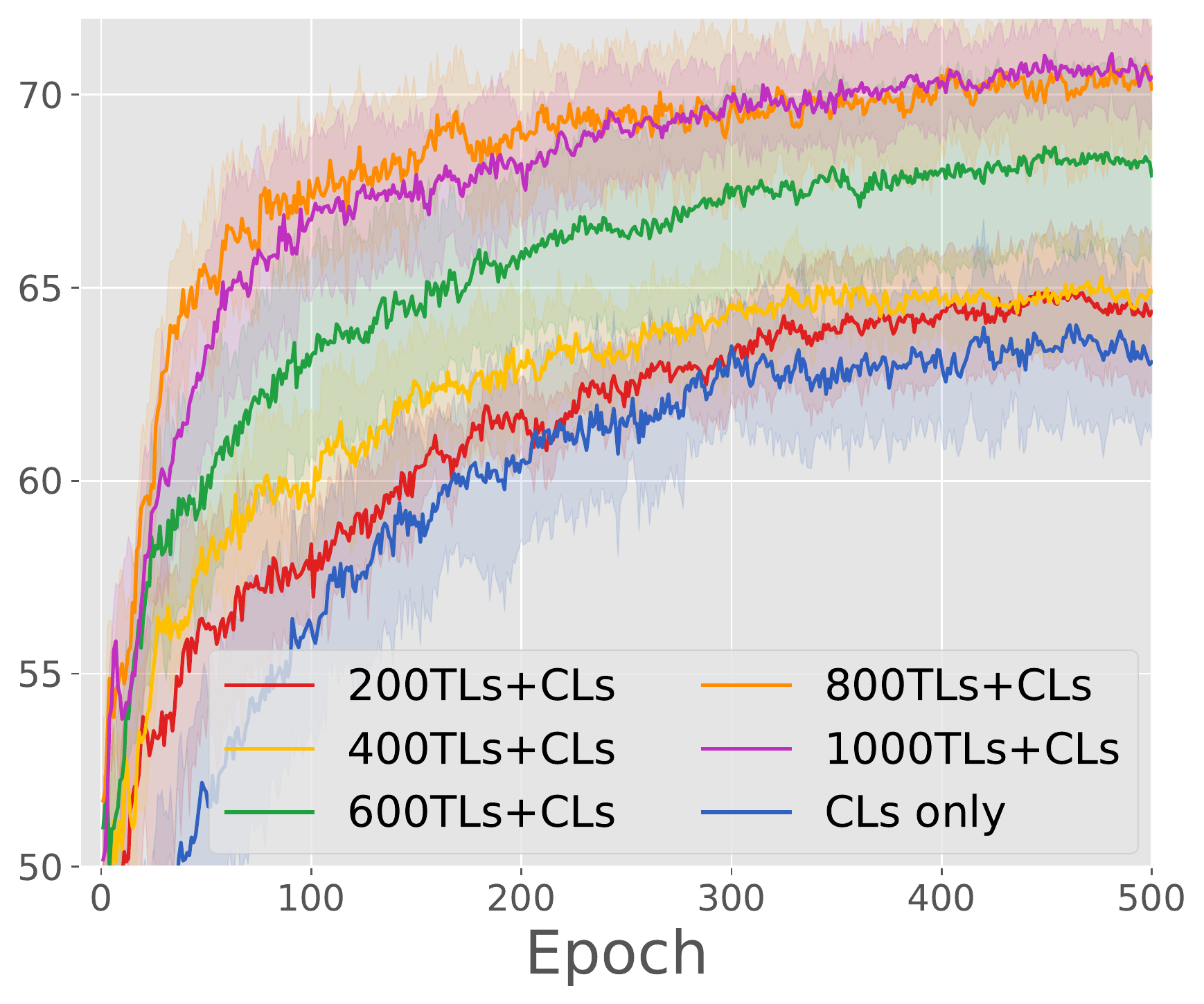}
}
%\subfigure[True Label \emph{vs.} Complementary Label (SVHN $\rightarrow$ MNIST).]
%{
%\includegraphics[width=0.23\textwidth]{truevscom.pdf}
%\includegraphics[width=0.23\textwidth]{epoch.pdf}
%}
%\vspace{-1em}
\caption{Test Accuracy vs. Epochs on $7$ CC-UDA Tasks in (a)-(g), and \emph{True Label} (TL) vs. \emph{Complementary Label} (CL) in (h). In (a)-(g), we compare the target-domain accuracy of one-step approach, i.e., CLARINET (\emph{ours}), with that of two-step approach (\emph{ours}). In (h), ``200TLs'' represents ordinary UDA method trained with $200$ true-label source data. ``200TLs+CLs'' means a CLARINET trained with $200$ true-label source data and complementary-label source data and ``CLs Only'' represents a CLARINET trained with complementary-label source data.} 
\label{fig:ccom}
% \vspace{-1em}
\end{figure*}

%a CDAN (a UDA method)
% ordinary UDA methods

\begin{table*}[t]
  \centering
  \small
  \caption{Results on $7$ CC-UDA Tasks. Bold value represents the highest accuracy (\%) on each row. Please note, the two-step methods and CLARINET are all first proposed in our paper. }
    \setlength{\tabcolsep}{7mm}{
    \begin{tabular}{lccccc}
    \toprule
    \multirow{2}[4]{*}{Tasks} & \multirow{2}[4]{*}{GAC} & \multicolumn{3}{c}{Two-step approaches (\emph{ours})} & \multirow{1}[4]{*}{CLARINET} \\
\cmidrule{3-5}          &       & GAC+DAN & GAC+DANN & GAC+CDAN\_E & (\emph{ours}) \\
    \midrule
    $C \rightarrow T$   & 45.167  & 45.711$\pm$0.535   & 45.628$\pm$0.572  & 45.228$\pm$0.270  & \textbf{47.083$\pm$1.395 } \\
    $U \rightarrow M$   & 51.860  & 60.692$\pm$1.300  & 77.580$\pm$0.770  & 71.498$\pm$1.077  & \textbf{83.692$\pm$0.928 } \\
    $M \rightarrow U$   & 77.796  & 87.215$\pm$0.603  & 88.688$\pm$1.280  & 92.366$\pm$0.365  & \textbf{94.538$\pm$0.292 } \\
    $S \rightarrow M$   & 39.260  & 45.132$\pm$1.363  & 50.882$\pm$2.440  & 61.922$\pm$2.983  & \textbf{63.070$\pm$1.990 }  \\
    $M \rightarrow m$   & 45.045  & 43.346$\pm$2.224  & 62.273$\pm$2.261  & 71.379$\pm$0.620  & \textbf{71.717$\pm$1.262 } \\
    $Y \rightarrow M$   & 77.070  & 81.150$\pm$0.591  & 92.328$\pm$0.138  & 95.532$\pm$0.873  & \textbf{97.040$\pm$0.212 } \\
    $Y \rightarrow S$   & 72.480  & 78.270$\pm$0.311  & 75.147$\pm$1.401  & 82.878$\pm$0.278  & \textbf{84.499$\pm$0.537 } \\
    \midrule
    Average  & 58.383  & 63.074   & 70.361  & 74.400  & \textbf{77.377 } \\
    \bottomrule
    \end{tabular}}%
  \label{tab:ccom}%
\end{table*}%

\subsection{Experimental Setup}
\label{Asec:Exp_set}

In general, we compose feature extractor $G$ from several CNN layers and one fully connected layer, picking their structures from previous works. The label predictor $F$ and domain discriminator $D$ all share the same structure in all tasks, following CDAN \cite{long2018conditional}. 

\begin{figure*}[!tp]
	\begin{center}
		{\includegraphics[width=0.92\textwidth]{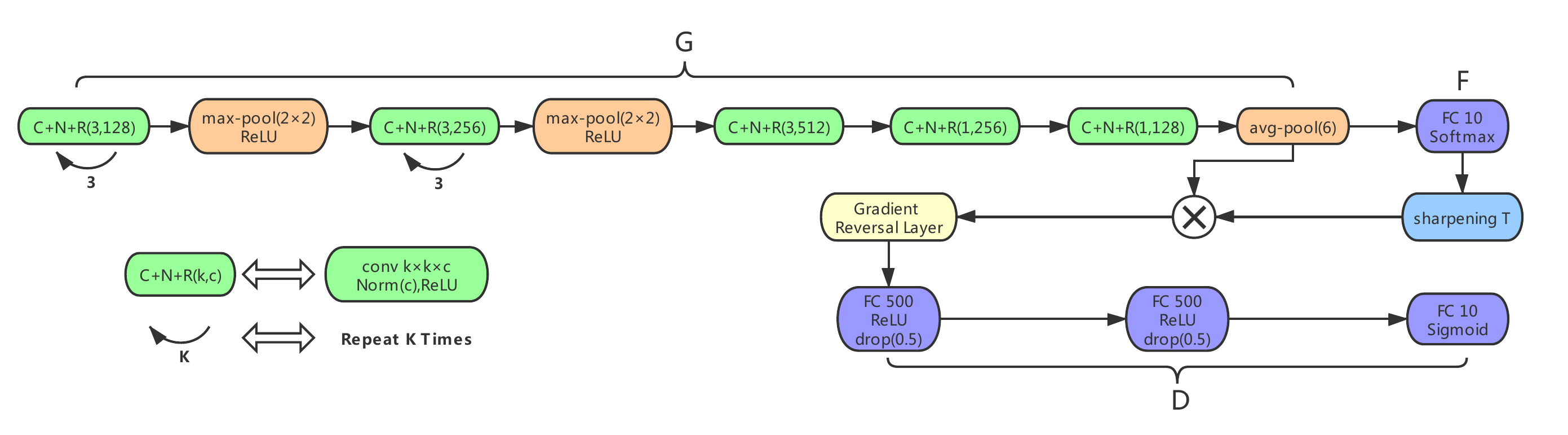}}
		\end{center}
		\caption{{The architecture of CLARINET for $C \rightarrow T$ task. Feature extractor $G$ is adopted from \protect\cite{DBLP:conf/iclr/}.}}
		\label{fig: CTNet}
	\vspace{-1em}
\end{figure*}

\begin{figure*}[!tp]
	\begin{center}
		{\includegraphics[width=0.75\textwidth]{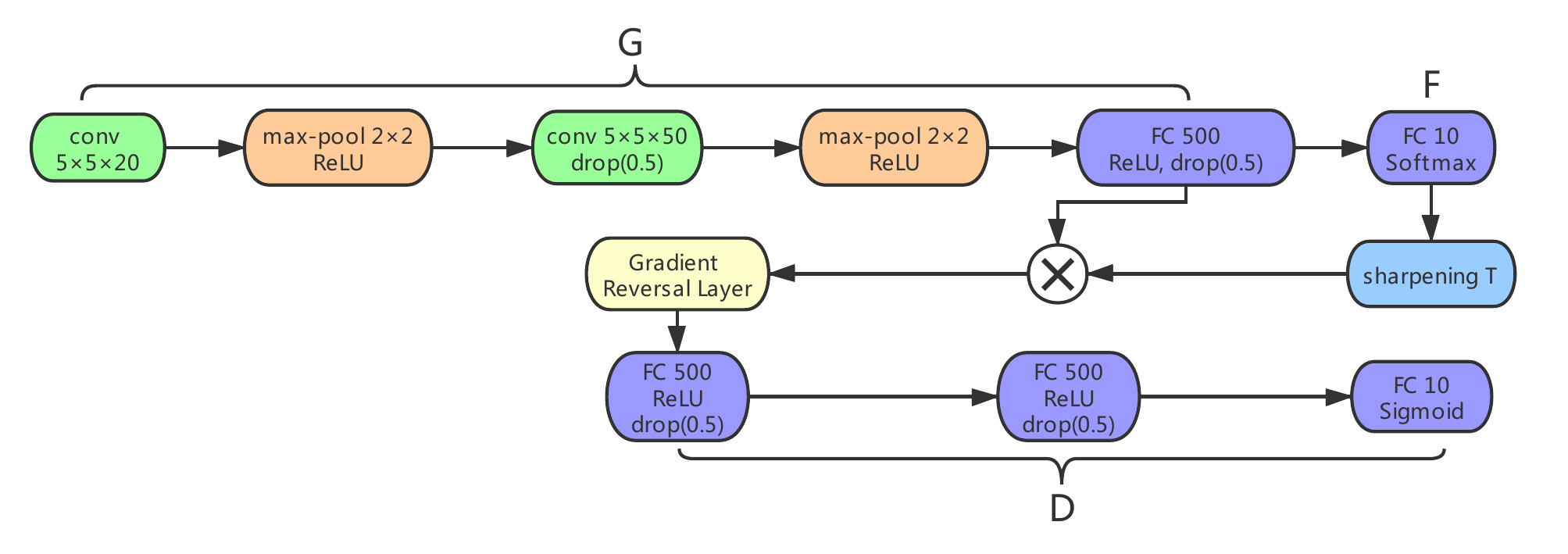}}
		\end{center}
		\caption{{The architecture of CLARINET for $U \rightarrow M$ and $M \rightarrow U$ tasks. Following \protect\cite{long2018conditional}.}}
		\label{fig: LeNet}
	\vspace{-1em}
\end{figure*}

\begin{figure*}[!tp]
	\begin{center}
		{\includegraphics[width=0.9\textwidth]{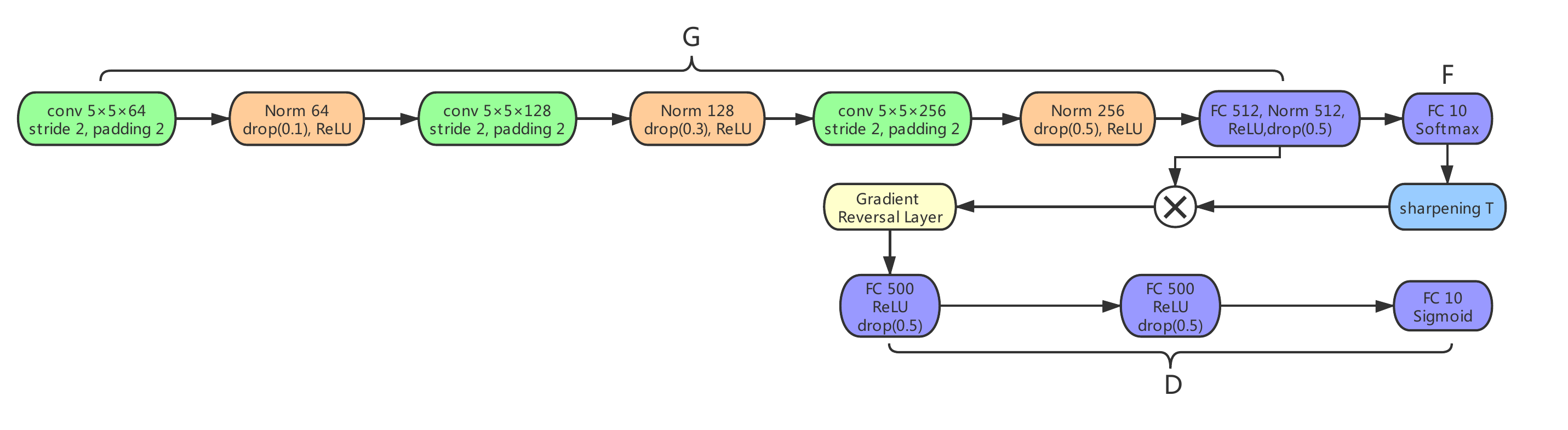}}
		\end{center}
		\caption{{The architecture of CLARINET for $S \rightarrow M$, $Y \rightarrow M$ and $Y \rightarrow S$ tasks. Following \protect\cite{long2018conditional}.}}
		\label{fig: DTN}
	\vspace{-1em}
\end{figure*}

\begin{figure*}[!tp]
	\begin{center}
		{\includegraphics[width=0.75\textwidth]{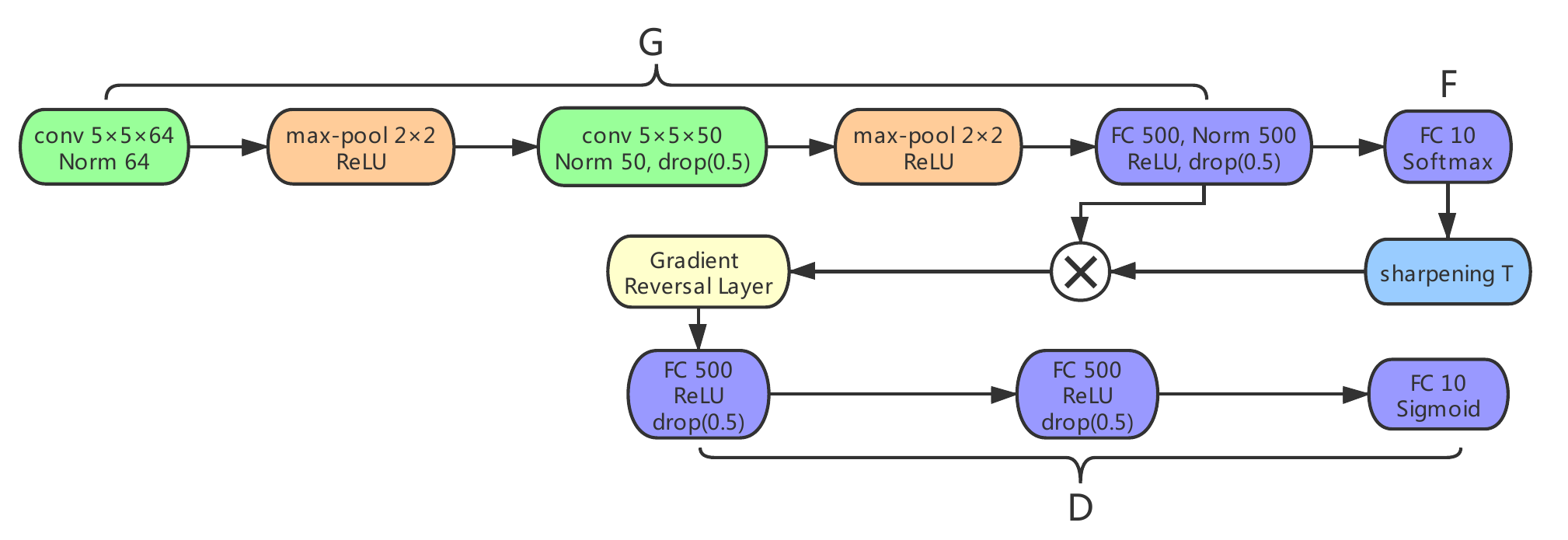}}
		\end{center}
		\caption{{The architecture of CLARINET for $M \rightarrow m$ task. Feature extractor $G$ is adopted from \protect\cite{ganin2016domain}.}}
		\label{fig: MmNet}
	\vspace{-1em}
\end{figure*}

More precisely, four different architectures of $G$ are used in our experiments (as shown in figures \ref{fig: CTNet}, \ref{fig: LeNet}, \ref{fig: DTN}, \ref{fig: MmNet}). In $C \rightarrow T$ task, we adopt the the structure provided in MT \cite{DBLP:conf/iclr/}. For $U \rightarrow M$ and $M \rightarrow U$ tasks, we use the LeNet provided in CDAN \cite{long2018conditional}. For $S \rightarrow M$, $Y \rightarrow M$ and $Y \rightarrow S$ tasks, we use the DTN provided in CDAN \cite{long2018conditional}. In $M \rightarrow m$ task, we adopt the the structure provided in DANN \cite{ganin2016domain}.  

%We use no more than $3$ layers of CNN and $2$ fully connected layers in the classifier and same $3$ fully connected layers in the domain discriminator.

We follow the standard protocols for unsupervised domain adaptation and compare the average classification accuracy based on $5$ random experiments. For each experiment, we take the result of the last epoch.

The batch size is set to $128$ and we train $500$ epochs. SGD optimizer (momentum= $0.9$, weight\_decay= $5e-5$) is with an initial learning rate of  $0.005$ in the adversarial network and $5e-5$ in the classifier. In sharpening function $T$, $l$ is set to $0.5$. For other special parameters in baselines, we all follow the original settings. We implement all methods with default parameters by PyTorch. The code of CLARINET is available at \httpsurl{github.com/Yiyang98/BFUDA}.

\begin{table*}[htbp]
  \centering
    \small
    \caption{Results on $7$ PC-UDA Tasks. Amount represents the number of true-label data in the source domain. In general, the accuracy of CLARINET increases when increasing the amount of true-label source data.}
    \setlength{\tabcolsep}{4.1mm}{
    \begin{tabular}{ccccccc}
    \toprule
    \multicolumn{7}{c}{Amount of of true-label source data} \\
    \midrule
    \multirow{2}[4]{*}{Tasks} & \multicolumn{2}{c}{0} & \multicolumn{2}{c}{200} & \multicolumn{2}{c}{400} \\
\cmidrule{2-7}          & true only & com only & true only & com+true & true only & com+true \\
    \midrule
    $C \rightarrow T$   & -     & 47.083$\pm$1.395 & 11.839$\pm$0.019 & 49.408$\pm$1.776 & 13.875$\pm$1.366 & 49.553$\pm$1.362 \\
    $U \rightarrow M$   & -     & 83.692$\pm$0.928 & 74.180$\pm$1.218 & 88.584$\pm$1.040 & 79.200$\pm$0.837 & 89.480$\pm$1.660 \\
    $M \rightarrow U$   & -     & 94.538$\pm$0.292 & 78.011$\pm$1.473 & 93.204$\pm$1.398 & 83.204$\pm$1.545 & 94.677$\pm$0.576 \\
    $S \rightarrow M$   & -     & 63.070$\pm$1.990 & 25.772$\pm$0.146 & 64.734$\pm$2.096 & 41.232$\pm$1.089 & 64.912$\pm$0.928 \\
    $M \rightarrow m$   & -     & 71.717$\pm$1.262 & 59.414$\pm$1.381 & 70.730$\pm$1.620 & 59.805$\pm$0.951 & 71.198$\pm$0.623 \\
    $Y \rightarrow M$   & -     & 97.040$\pm$0.212 & 49.232$\pm$1.354 &      97.182$\pm$0.383 & 60.640$\pm$1.570 &
    97.242$\pm$0.117 \\
    $Y \rightarrow S$   & -     & 84.499$\pm$0.537 & 23.009$\pm$1.102 & 84.269$\pm$0.814 & 49.120$\pm$1.236 & 85.538$\pm$0.596 \\
    \midrule
    Average & -     & 77.377  & 45.922  & 78.302  & 55.297  & 78.943  \\
    \midrule
    \multirow{2}[4]{*}{Tasks} & \multicolumn{2}{c}{600} & \multicolumn{2}{c}{800} & \multicolumn{2}{c}{1000} \\
\cmidrule{2-7}          & true only & com+true & true only & com+true & true only & com+true \\
    \midrule
    $C \rightarrow T$   & 17.722$\pm$2.626 & 50.897$\pm$0.969 & 19.278$\pm$0.853 & 51.058$\pm$1.737 & 20.972$\pm$1.061 & 53.297$\pm$1.655 \\
    $U \rightarrow M$   & 82.532$\pm$0.859 & 90.358$\pm$1.938 & 85.800$\pm$0.621 & 91.106$\pm$0.561 & 88.184$\pm$1.280 & 93.342$\pm$1.294 \\
    $M \rightarrow U$   & 83.925$\pm$1.511 & 94.839$\pm$0.254 & 85.839$\pm$2.074 & 94.796$\pm$0.104 & 85.699$\pm$0.777 & 95.022$\pm$0.280 \\
    $S \rightarrow M$   & 41.680$\pm$0.525 & 67.898$\pm$1.625 & 51.652$\pm$0.850 & 70.416$\pm$1.819 & 53.500$\pm$1.872 & 70.446$\pm$1.358 \\
    $M \rightarrow m$   & 63.757$\pm$1.344 & 72.732$\pm$0.947 & 65.161$\pm$0.766 & 73.050$\pm$1.264 & 68.522$\pm$1.285 & 73.336$\pm$0.727 \\
    $Y \rightarrow M$   & 76.802$\pm$1.649 &   97.178$\pm$0.396 & 85.286$\pm$1.363 &      96.842$\pm$0.267 & 86.470$\pm$1.646 &
    96.948$\pm$0.266 \\
    $Y \rightarrow S$   & 67.922$\pm$1.079 &   85.921$\pm$1.098 & 67.788$\pm$1.878 &      86.772$\pm$0.291 & 74.654$\pm$1.054 &
    87.024$\pm$0.542 \\
    \midrule
    Average & 62.049  & 79.975  & 65.829  & 80.577  & 68.286  & 81.345  \\
    \bottomrule
    \end{tabular}}%
  \label{tab:pcom}%
\end{table*}%
%bottomrule

% Table generated by Excel2LaTeX from sheet 'Sheet2'
\begin{table*}[htbp]
  \centering
  \small
  \caption{Ablation Study. Bold value represents the highest accuracy (\%) on each column. Obviously to see, UDA methods cannot handle complementary-label based UDA tasks directly. We also prove that the conditioning adversarial part and the sharpening function $T$ can help improve the adaptation performance.}
  \setlength{\tabcolsep}{0.5mm}{
    \begin{tabular}{lcccccccc}
    \toprule
    Methods & $C \rightarrow T$   & $U \rightarrow M$   & $M \rightarrow U$   & $S \rightarrow M$   & $M \rightarrow m$   & $Y \rightarrow M$   & $Y \rightarrow S$   & Average \\
    \midrule
    C w/ $L_{CE}$   & 6.481$\pm$2.536      & 0.455$\pm$0.722  & 0.055$\pm$0.129  & 3.708$\pm$0.688  & 7.088$\pm$0.424  & 1.832$\pm$0.102  & 1.298$\pm$0.070  & 2.987  \\
    C w/o $c$ & 41.908$\pm$2.796  & \textbf{ 84.302$\pm$1.127 } & 93.301$\pm$0.465  & 44.500$\pm$2.088  & 70.994$\pm$0.749  & 94.382$\pm$0.150  & 83.408$\pm$0.545  & 73.256  \\
    C w/o $T$ & 43.075$\pm$2.553  & 83.192$\pm$1.796  & 93.419$\pm$0.588  & 52.438$\pm$1.927  & \textbf{ 72.128$\pm$1.569 } & 95.442$\pm$1.004  & 83.055$\pm$0.652  & 74.678  \\
    CLARINET & \textbf{ 47.083$\pm$1.395 } & 83.692$\pm$0.928  & \textbf{ 94.538$\pm$0.292 } & \textbf{ 63.070$\pm$1.990 } & 71.717$\pm$1.262  & \textbf{ 97.040$\pm$0.212 } & \textbf{ 84.499$\pm$0.537 } & \textbf{ 77.377 } \\
    \bottomrule
    \end{tabular}}%
  \label{tab:ablation}%
\end{table*}%

\iffalse
\begin{figure*}[t]
\centering
\subfigure
{
\includegraphics[width=0.38\textwidth]{truevscom2.pdf}
%\includegraphics[width=0.4\textwidth]{epoch.pdf}
}
\subfigure
{
\includegraphics[width=0.38\textwidth]{epoch2.pdf}
}
\vspace{-1em}
\caption{True Label (TL) \emph{vs.} Complementary Label (CL) in SVHN $\rightarrow$ MNIST task. ``200TLs'' represents a CDAN (a UDA method) trained with $200$ true-label source data. ``200TLs+CLs'' means a CLARINET trained with $200$ true-label source data and complementary-label source data and ``CLs Only'' represents a CLARINET trained with complementary-label source data.} 
\label{fig:pcom}
% \vspace{-1em}
\end{figure*}
\fi

\subsection{Results on CC-UDA Tasks}

Table \ref{tab:ccom} reports the target-domain accuracy of $5$ methods on $7$ CC-UDA tasks. As can be seen, our CLARINET performs best on each task and the average accuracy of CLARINET is significantly higher than those of baselines. Compared with GAC method, CLARINET successfully transfers knowledge from complementary-label source data to unlabeled target data. Since CDAN has shown much better adaptation performance than DANN and DAN \cite{long2018conditional}, GAC+CDAN should outperform other two-step methods on each task. However, on the task $U$$\rightarrow$$ M$, the accuracy of GAC+CDAN is much lower than that of GAC+DANN. This abnormal phenomenon shows that the noise contained in pseudo-label source data significantly reduces transferability of existing UDA methods. Namely, we cannot obtain the reliable adaptation performance by using two-step CC-UDA approach.

\textbf{$\emph{CIFAR} \rightarrow \emph{STL}$.} CIFAR and STL are 10-class object recognition datasets with colored images. We remove the non-overlapping classes (“frog” and “monkey”) and readjust the labels to align the two datasets. Namely this task reduce to a $9$-class classification problem. Furthermore, we downscale the $96 \times 96$ image dimesion of STL to match the $32 \times 32$ dimension of CIFAR. As shown in Figure \ref{fig:ccom} (a), two-step methods could hardly realize knowledge transfer, while our CLARINET’s performance surpasses others by a comfortable margin.

\textbf{$\emph{MNIST} \leftrightarrow \emph{USPS}$.} MNIST and USPS are both grayscale digits images, thus the distribution discrepancy between the two tasks is relatively small. As shown in Figure \ref{fig:ccom} (b) and (c), in both adaptation directions, CARINET all achieve the best performance far above other baselines.

\textbf{$\emph{SVHN} \rightarrow \emph{MNIST}$.} SVHN and MNIST are both digit datasets. Whereas MNIST consists of black-and-white handwritten digits, SVHN consists of crops of colored,
street house numbers. MNIST has a lower image dimensionality than SVHN, thus we adopt the dimension of MNIST to $32 \times 32$ with three channels to match SVHN. Because of the above factors, the gap between two distributions are relatively larger compared to that of the $\emph{MNIST} \leftrightarrow \emph{USPS}$. As shown in Figure \ref{fig:ccom} (d), GAC+CDAN perform much better than GAC+DAN and GAC+DANN, but still worse than our CLARINET.

\textbf{$\emph{MNIST} \rightarrow \emph{MNIST-M}$.} MNIST-M is a transformed dataset from MNIST, which was composed
by merging clips of a background from the BSDS500 datasets \cite{arbelaez2010contour}. For a human the classification task on MNIST-M only becomes slightly harder, whereas for a CNN network trained on MNIST, this domain is quite different, as the background
and the strokes are no longer constant. As shown in Figure \ref{fig:ccom} (e), Our method is slightly more effective than GAC+CDAN and far more effective than the other two methods. 

\textbf{$\emph{SYN-DIGITS} \rightarrow \emph{MNIST}$.} This adaptation reflects a common adaptation problem of transferring from synthetic images to real images. The SYN-DIGITS dataset consists of a huge amount of data, generated from Windows fonts by varying the text, positioning, orientation, background,
stroke color, and the amount of blur. As shown in Figure \ref{fig:ccom} (f), our method outperforms other baselines and achieves pretty high accuracy. Thus with sufficient source data, CLARINE could achieve excellent results.

\textbf{$\emph{SYN-DIGITS}  \rightarrow \emph{SVHN}$.} This adaptation is another common adaptation problem of transferring from synthetic images to real images, but is more challenging than in the case of the MNIST
experiment. As shown in Figure \ref{fig:ccom} (g), our method is obviously more effective than other baselines. GAC+DANN does not apply to this task, achieving the lowest accuracy. 
%even worse than GAC+DAN.

%Figure \ref{fig:ccom} ((a)-(f)) shows the target-domain accuracy of two-step methods and CLARINET on $7$ BFUDA tasks when increasing epochs. It is clear to see that the accuracy of CLARINET gradually and steadily increases and eventually converges, achieving the best accuracy on most tasks. The accuracy of GAC+CDAN always reaches plateau quickly. For GAC+DANN, its accuracy is unstable and significantly drops after certain epochs on $M$$\rightarrow$$ U$, $M$$\rightarrow$$ m$ tasks. While the test accuracy of GAC+DAN is relatively stable but not satisfactory.

\subsection{Results on PC-UDA Tasks} 
\label{true ex}

Table \ref{tab:pcom} reports the target-domain accuracy of CLARINET on PC-UDA tasks when we have different amount of true-label source data. ``true only'' means training on a certain number of true-label source data with ordinary UDA method. ``com only'' means training on complementary-label source data only with CLARINET. ``com+true'' stands for training on a certain number of true-label source data and complementary-label source data with CLARINET. In general, the accuracy of CLARINET increases when increasing the amount of true-label source data from $0$ to $1000$. Thus, it is proved that CLARINET can sufficiently leverage true-label source data to improve adaptation performance. 

The improvement is especially evident on $U$$\rightarrow$$ M$ task and $S$$\rightarrow$$ M$ task. For $U$$\rightarrow$$ M$ task, this is probably because the dataset sample size of USPS is relatively small, true-label data actually has occupied a large part. For $S$$\rightarrow$$ M$ task, SVHN is complicated for complementary-label learning. Hence adding a small number of true-label data could help to train a more accurate classifier. This phenomenon also reminds us that for complex datasets, adding some 
true-label data to assist training would be pretty appropriate. On $Y$$\rightarrow$$ M$ task, adding true-label source data does not bring significant improvement, which is most likely due to the result on complementary-label data is already relatively good and true-label source data is unable to assist in achieving better result.
%We also observe that the improvement on the accuracy of CLARINET on $U$$\rightarrow$$M$, $S$$\rightarrow$$ M$ and $Y$$\rightarrow$$ S$ tasks is more than that on other tasks when increasing the ratio.

We also compare the efficacy of true-label source data with complementary-label source data. Taking $S$$\rightarrow$$ M$ task as an example (shown in the left part of Figure \ref{fig:ccom} (h)), we compare the target-domain accuracy of ordinary UDA method trained with different amount of true-label source data and that of CLARINET trained with complementary-label source data only (``CLs Only''). The accuracy decreases significantly when reducing the amount of true-label source data, which suggests that sufficient true-label source data are inevitably required in UDA scenario. Then we compare the target-domain accuracy of CLARINET trained with complementary-label source data only with that of CLARINET trained with different amount of true-label and complementary-label source data. It is clear that CLARINET effectively uses two kinds of data to obtain better adaptation performance than using complementary-label source data only. Besides, as the number of true-label source data used increases, the classification accuracy becomes better (shown in the right part of Figure \ref{fig:ccom} (h)).

\subsection{Ablation Study}

Finally, we conduct experiments to show the contributions of different components in CLARINET. We consider following baselines: 

\begin{itemize}
\item C w/ $L_{CE}$: train \underline{C}LARINET by Algorithm~\ref{alg:algorithm}, while {replacing} $\overline{L}_s(G,F,\overline{D}_s)$ by \underline{c}ross-\underline{e}ntropy loss.
\item C w/o $c$ : train \underline{C}LARINET \underline{without} \underline{c}onditioning, namely train the domain discriminator $D$ only based on feature representations $g_s$ and  $g_t$.   
\item C w/o $T$: train \underline{C}LARINET by Algorithm~\ref{alg:algorithm}, \underline{without} sharpening function $\underline{T}$.
\end{itemize}

C w/ $L_{CE}$ uses the cross-entropy loss to take place of complementary-label loss. Actually, it stands for applying ordinary UDA methods directly on complementary-label based UDA tasks. The target-domain accuracy of C w/ $L_{CE}$ will show whether UDA methods can address the complementary-label based UDA problem. C w/o $c$ train the domain discriminator $D$ only based on feature representations $g_s$ and $g_t$, thus the result could indicate whether the conditional adversarial way could capture the multimodal structures so as to improve the transfer effect. Please notice, the sharpening function $T$ is useless in this network as it works on the label prediction $f_s$ and $f_t$. Comparing CLARINET with C w/o $T$ reveals if the sharpening function $T$ takes effect. 

As shown in Table~\ref{tab:ablation}, the target-domain accuracy of C w/ $L_{CE}$ is much lower than that of other methods. Namely, UDA methods cannot handle complementary-label based UDA tasks directly. Its result is not even as good as random classification, as the network is trained taking the wrong label as the target result. Comparing with C w/o $T$, C w/o $c$ has a worse performance, which proves that the conditional adversarial way could really improve the transfer effect. Therefore, it is necessary to capture the multimodal structures of distributions with cross-covariance dependency between the features and classes in the field of adversarial based UDA. Although C w/o $T$ achieves better accuracy than other baselines, its accuracy still worse than CLARINET's. The result reveals that the sharpening function $T$ helps to capture multimodal structures of distributions on basis of the characteristics of complementary-label learning. Thus, the sharpening function $T$ can improve the adaptation performance. 

\section{Conclusion and Further Study}

This paper presents a new setting, complementary-label based UDA, which exploits economical complementary-label source data instead of expensive true-label source data. We consider two cases of the complementary-label based UDA, one is that the source domain only contains complementary-label data (CC-UDA), and the other is that the source domain has plenty of complementary-label data and a small amount of true-label data (PC-UDA). Since existing UDA methods cannot address complementary-label based UDA problem, we propose a novel, one-step approach, called \emph{\underline{c}omplementary \underline{l}abel advers\underline{ari}al \underline{net}work} (CLARINET). CLARINET could handle both CC-UDA and PC-UDA tasks. Experiments conducted on $7$ complementary-label based UDA tasks confirm that CLARINET effectively achieves distributional adaptation from complementary-label source data to unlabeled target data and outperforms a series of competitive baselines. In the future, we plan to explore more effective ways to solve complementary-label based UDA and extend the application of complementary labels in domain adaptation.

\section*{Acknowledgements}

The work presented in this paper was supported by the Australian Research Council (ARC) under FL190100149. The first author particularly thanks the support by UTS-CAI during her visit.

\ifCLASSOPTIONcaptionsoff
  \newpage
\fi



\bibliographystyle{IEEEtran}
\bibliography{main}

%\begin{thebibliography}{1}


%\end{thebibliography}


\begin{IEEEbiography}[{\includegraphics[width=1in,height=1.25in,clip,keepaspectratio]{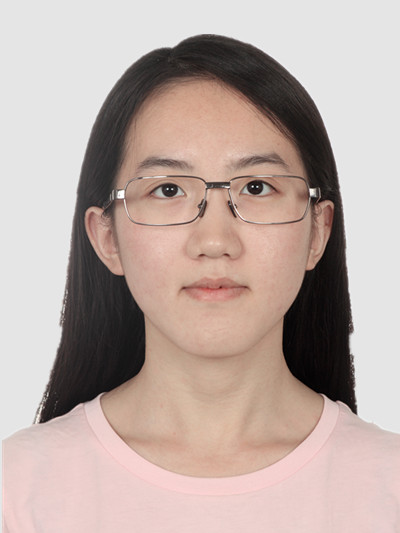}}]{Yiyang Zhang} received her B.E. degree in Automation from Tsinghua University, P.R. China, in 2018. She is currently a Master student in Data Science at Shenzhen International Graduate School, Tsinghua University, and a visiting student with the Centre for Artificial Intelligence (CAI), the University of Technology, Sydney (UTS). Her research interests include transfer learning and domain adaptation.
\end{IEEEbiography}

\begin{IEEEbiography}[{\includegraphics[width=1in,height=1.25in,clip,keepaspectratio]{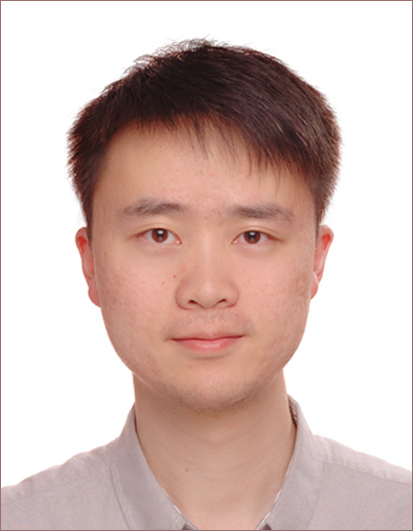}}]{Feng Liu}
is a Doctoral candidate in Centre for Artificial intelligence, Faculty of Engineering and Information Technology, University of Technology Sydney, Australia. He received an M.Sc. degree in probability and statistics and a B.Sc. degree in pure mathematics from the School of Mathematics and Statistics, Lanzhou University, China, in 2015 and 2013, respectively. His research interests include domain adaptation and two-sample test. He has served as a senior program committee member for ECAI and program committee members for NeurIPS, ICML, IJCAI, CIKM, FUZZ-IEEE, IJCNN and ISKE. He also serves as reviewers for TPAMI, TNNLS, TFS and TCYB. He has received the UTS-FEIT HDR Research Excellence Award (2019), Best Student Paper Award of FUZZ-IEEE (2019) and UTS Research Publication Award (2018).
\end{IEEEbiography}

\begin{IEEEbiography}[{\includegraphics[width=1.0in,height=1.8in,clip,keepaspectratio]{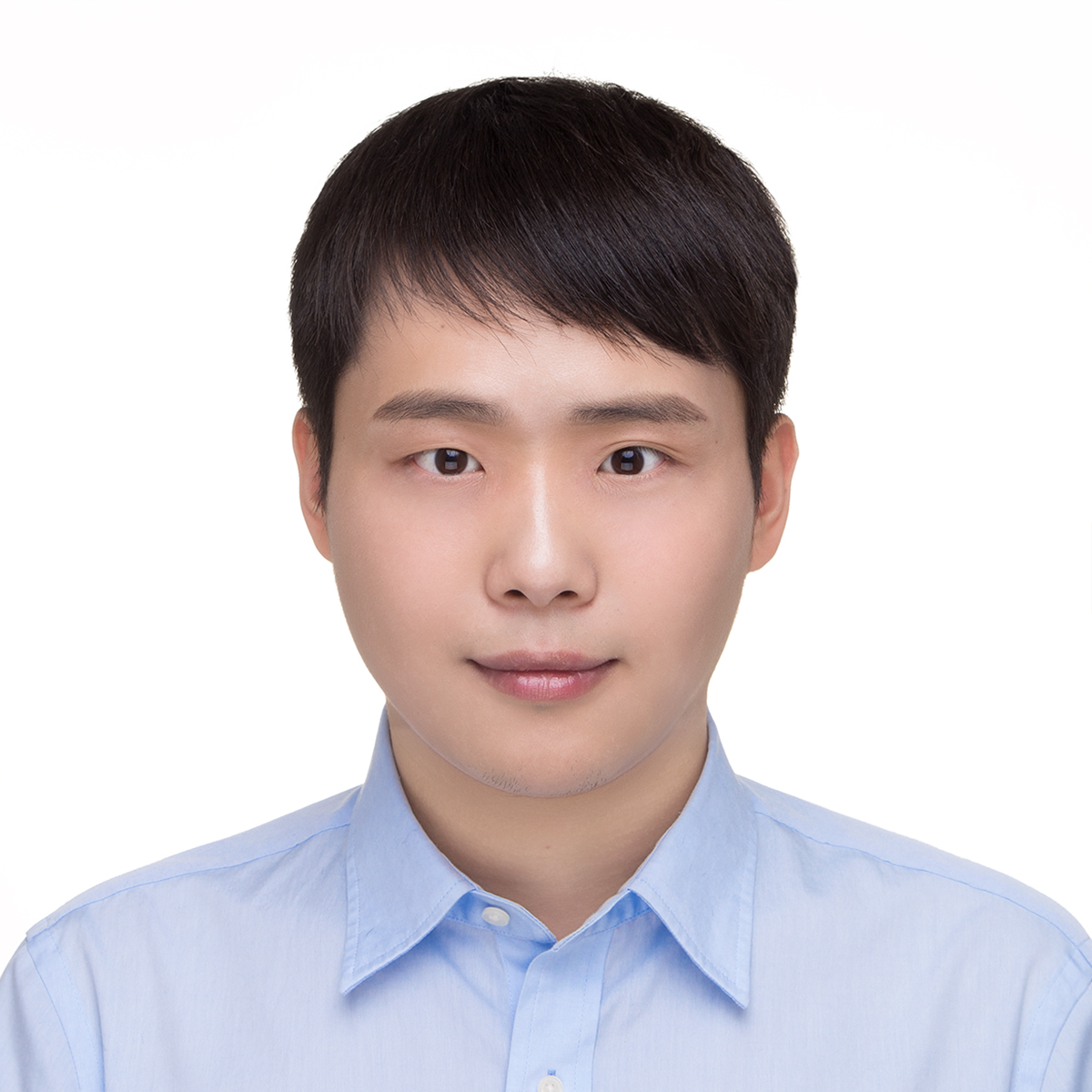}}]{Zhen Fang}
received his M.Sc. degree in pure mathematics from the School of Mathematical Sciences Xiamen University, Xiamen, China, in 2017. He is working toward a PhD degree with the Faculty of Engineering and Information Technology, University of Technology Sydney, Australia. His research interests include transfer learning and domain adaptation. He is a Member of the Decision Systems and e-Service Intelligence (DeSI) Research Laboratory, CAI, University of Technology Sydney.
\end{IEEEbiography}

\begin{IEEEbiography}[{\includegraphics[width=1in,height=1.25in,clip,keepaspectratio]{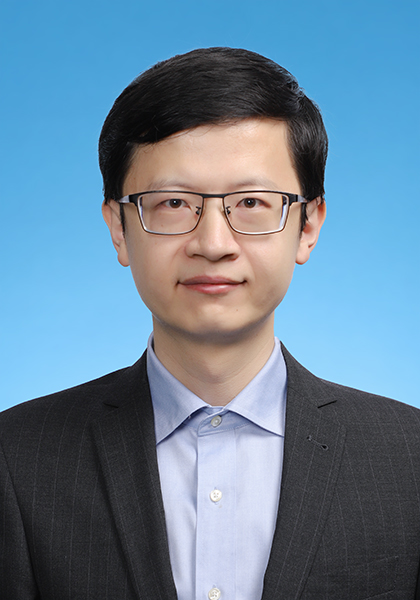}}]{Bo Yuan} received the B.E. degree from Nanjing University of Science and Technology, P.R.China, in 1998, and the M.Sc. and Ph.D. degrees from The University of Queensland
(UQ), Australia, in 2002 and 2006, respectively, all in Computer Science. From 2006 to
2007, he was a Research Officer on a project funded by the Australian Research Council at UQ. He is currently an Associate Professor in the Division of Informatics, Shenzhen International Graduate School, Tsinghua University, and a member of the Intelligent Computing Lab. His research interests include Intelligent Computing and Pattern Recognition.
\end{IEEEbiography}

\begin{IEEEbiography}[{\includegraphics[width=1in,height=1.25in,clip,keepaspectratio]{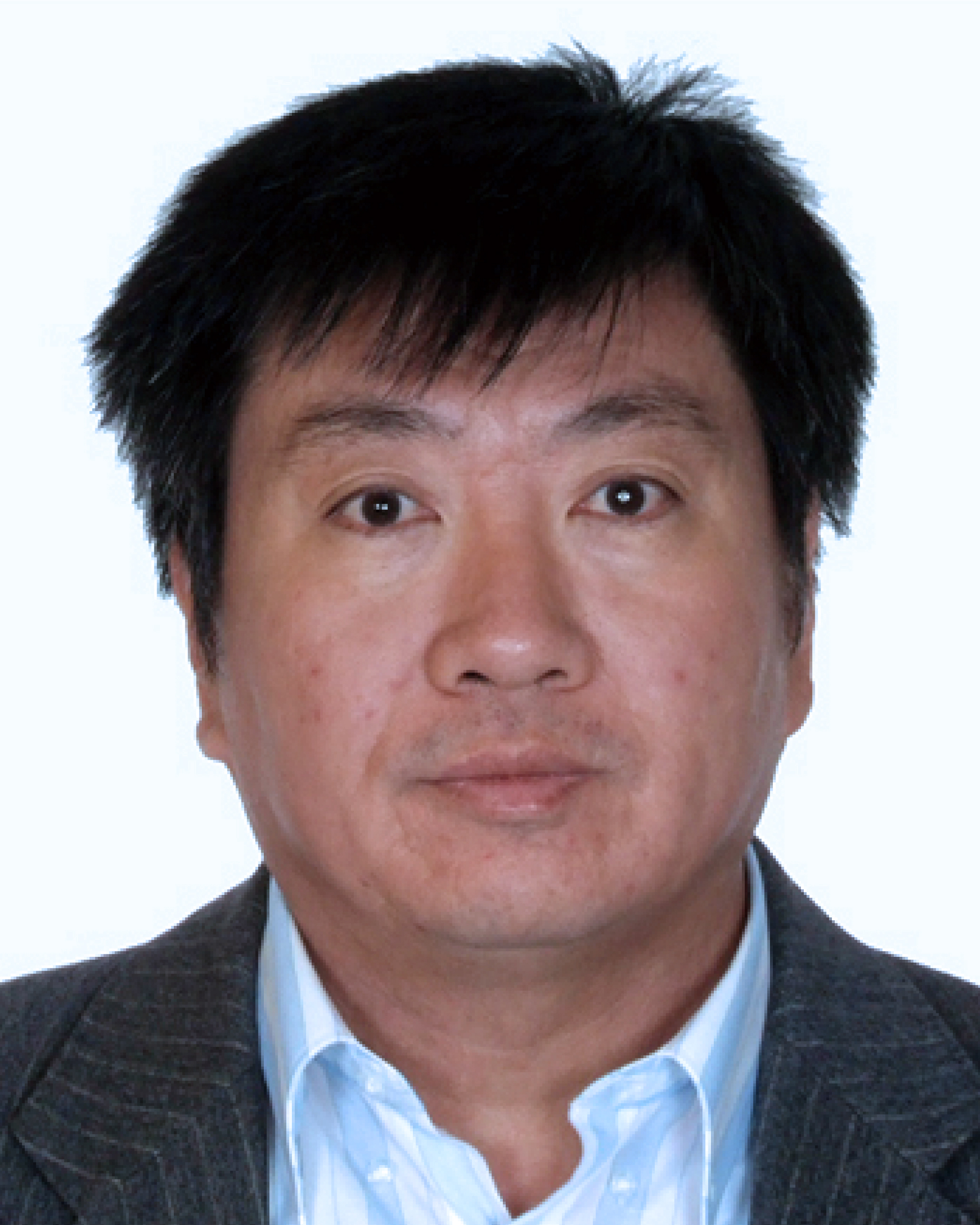}}]{Guangquan Zhang}
is a Professor and Director of the Decision Systems and e-Service Intelligent (DeSI) Research Laboratory, Faculty of Engineering and Information Technology, University of Technology Sydney, Australia. He received his PhD in applied mathematics from Curtin University of Technology, Australia, in 2001.
His research interests include fuzzy machine learning, fuzzy optimization, and machine learning and data analytics. He has authored four monographs, five textbooks, and 350 papers including 160 refereed international journal papers. Dr. Zhang has won seven Australian Research Council (ARC) Discovery Project grants and many other research grants. He was awarded an ARC QEII Fellowship in 2005. He has served as a member of the editorial boards of several international journals, as a guest editor of eight special issues for IEEE Transactions and other international journals, and has co-chaired several international conferences and work-shops in the area of fuzzy decision-making and knowledge engineering. 
\end{IEEEbiography}

\begin{IEEEbiography}[{\includegraphics[width=1in,height=1.25in,clip,keepaspectratio]{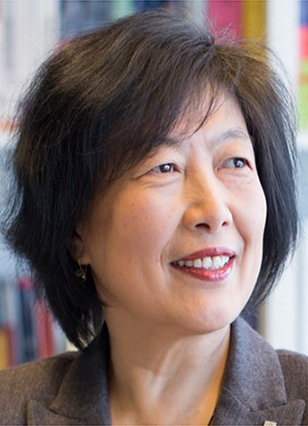}}]{Jie Lu} (F’18) is a Distinguished Professor and the Director of the Centre for Artificial Intelligence at the University of Technology Sydney, Australia. She received her PhD degree from Curtin University of Technology, Australia, in 2000. Her main research interests arein the areas of fuzzy transfer learning, concept drift, decision support systems, and recommender systems. She is an IEEE fellow, IFSA fellow and Australian Laureate fellow. She has published six research books and over 450 papers in refereed journals and conference proceedings; has won over 20 ARC Laureate, ARC Discovery Projects, government and industry projects. She serves as Editor-In-Chief for Knowledge-Based Systems (Elsevier) and Editor-In-Chief for International journal of computational intelligence systems. She has delivered over 25 keynote speeches at international conferences and chaired 15 international conferences. She has received various awards such as the UTS Medal for Research and Teaching Integration (2010), the UTS Medal for Research Excellence (2019), the Computer Journal Wilkes Award (2018), the IEEE Transactions on Fuzzy Systems Outstanding Paper Award (2019), and the Australian Most Innovative Engineer Award (2019).
\end{IEEEbiography}

\end{document}